\documentclass[runningheads]{llncs}
\usepackage{hyperref}
\usepackage{graphicx}
\usepackage{amsmath}
\graphicspath{{./Graphs_for_paper/}{./Figures_paper/}{./Final_figures/}}
\usepackage{caption}
\usepackage{subcaption}
\usepackage{amssymb}
\usepackage{multirow}
\usepackage{todonotes}
\usepackage{booktabs}
\usepackage{scrextend}
\usepackage{footnote}
\usepackage{csquotes}
\makesavenoteenv{tabular}
\makesavenoteenv{table}

\begin{document}
\title{Training Deep Neural Networks Without Batch Normalization}
\author{Divya Gaur\inst{1} \and
Joachim Folz\inst{2} \and
Andreas Dengel\inst{3}}
\authorrunning{Gaur, Divya et al.}
\institute{\email{
gaur@rhrk.uni-kl.de}\\
Technische Universit\"at Kaiserslautern, Gottlieb-Daimler-Strasse 47, 67663, Kaiserslautern, Germany
\\
\url{https://www.uni-kl.de}
\and
\email{joachim.folz@dfki.de}\\
\and
\email{andreas.dengel@dfki.de}\\
Deutsches Forschungszentrum für Künstliche Intelligenz (DFKI), Trippstadter Str. 122, 67663 Kaiserslautern
\url{https://www.dfki.de}
}
\maketitle              % typeset the header of the contribution
\begin{abstract}
Training neural networks is an optimization problem, and finding a decent set of parameters through gradient descent can be a difficult task.
A host of techniques has been developed to aid this process before and during the training phase.
One of the most important and widely used class of method is normalization.
It is generally favorable for neurons to receive inputs that are distributed with zero mean and unit variance, so we use statistics about dataset to normalize them before the first layer.
However, this property cannot be guaranteed for the intermediate activations inside the network.
A widely used method to enforce this property inside the network is batch normalization.
It was developed to combat covariate shift inside networks.
Empirically it is known to work, but there is a lack of theoretical understanding about its effectiveness and potential drawbacks it might have when used in practice.
This work studies batch normalization in detail, while comparing it with other methods such as weight normalization, gradient clipping and dropout.
The main purpose of this work is to determine if it is possible to train networks effectively when batch normalization is removed through adaption of the training process. 

\end{abstract}
\section{Introduction}
The idea of neural networks can be traced back to the 1940's, when the first related paper \cite{firstEverNNPaper} was published, which described the functioning of neurons in the brain through electrical circuits.
Most of the models developed around this time were a combination of mathematics and computational logic, but unfortunately due to lack of computational and storage capacity this development in artificial intelligence saw a decline, which is otherwise known as the first winter of AI.
The early 1980's again witnessed a rise in research related to neural networks in particular, with \cite{secondEverNNPaper} being one of the most prominent papers in this field, followed by other relevant research such as \cite{thirdEverNNPaper} which was published in the late 1990's.
By 2000 development of faster processors and GPUs offered a better landscape for training neural networks.
This advancement resulted in works like \cite{oh2004gpu}, \cite{secondGPU} where they used graphic hardware to speed up the computation.
This was followed by a rapid increase in research activity related to gradient-based methods.
But there were other unaddressed problems related to training deeper and relatively complex networks.
One major issue faced by computational models was the problem of vanishing gradients, where deeper layers were not able to learn anything and network performance solely depended on the choice of activation functions. 
The simplest and widely used solution for this was to pre-train each layer before concatenating it to form an actual deep neural network.
The initial networks which became very popular are AlexNet \cite{alexNet} and VGG \cite{VGG}.
All these networks faced similar problems that training deeper layers was more difficult and even shallow networks required careful selection of parameters and hyper-parameters.
Later, architectures such as ResNet \cite{resNetoriginal} were developed which use identity mapping, otherwise known as residual or skip connections, helping the networks to become deeper.
Though the ability to create much deeper networks up to 1000 layers deep was attributed to these skip connections, even the authors of ResNet \cite{resNetPaper} point out that introduction of batch normalization in \textit{residual networks} helps alleviate the inherent problems faced while optimizing deep networks.

Batch Normalization (BN) was first introduced in \cite{batchNormOriginal}.
The primary purpose of this work was to address the difficulties faced while training deeper neural networks.
The idea is to normalize the change activation of the inputs for layers, introduced as the \textit{Internal Covariate Shift} (ICS) in the original work.
This kind of conditioning the inputs was formerly introduced for single layer network introduced in \cite{leCunnWhitneningPaper}.
BN takes this idea and applies it to all the intermediate inputs for the layers inside network.
BN works by normalizing the parameters over a mini-batch rather than individual sample.
As the original work states, there are several advantages of carrying out the calculations on a mini-batch.
Firstly, the loss gradient would be much more reliable for a batch rather than single values. 
Additionally, it reduces computational effort and enables better usage of parallel computing infrastructure. 
Showing that there is an inherent dependence on the data distribution in each individual mini-batch.

Even though the theory of how it benefits the training of deep neural networks is so far not well understood, BN is an essential part of most modern network architectures.
Unless it causes serious problems the inclusion of BN is rarely questioned. \cite{BN_in_GAN}~demonstrates that for unsupervised tasks such as training Generative Adversarial Networks (GANs) using BN might introduce some instability in the trained model.
\cite{fix_BN_layer}~suggests that for object detection tasks where the mini-batch size has to be kept small it is useful to use \textit{fixed} BN layers, i.e., they are pre-trained on a different dataset which allows it to be trained properly, and the trained statistics are then used for working with other models helping them train faster and train well.
In addition to this, BN has poor compatibility with multi-GPU training, where workarounds have to be used to achieve expected training results.
With typical data-parallel training (each GPU handles a portion of the batch) only partial batch statistics are used, which introduces small deviations in model behavior across GPUs.
Alternatively batch statistics can be synchronized between GPUs for each BN layer, which reduces performance.
Despite these shortcoming, the prevalence of BN in published work shows that researchers currently prefer to work around them rather than replacing BN.
Hence, the remainder of this work attempts to highlight how BN affects training and its relation with ICS as a topic of current research.
In particular experiments will be conducted to answer the following questions: 

\begin{enumerate}

\item \textit{Is it possible to achieve similar performance without BN if gradient magnitudes are maintained in a manner where they neither diminish nor grow out of bounds?}

\item \textit{Can large learning rates be used without BN and without compromising the performance of the trained model?}

\item \textit{Is it possible to achieve similar performance by replacing BN layers with less complex methods such as weight normalized convolution layers~\cite{weightNorm}, gradient clipping~\cite{gradientClipping} and dropout~\cite{dropoutPaper}?}

\end{enumerate}

The relevant work which makes the basis for this paper is discussed in the section below, followed by experiments and discussion/conclusion sections.

\section{Related Work}

When BN was first introduced, the authors claimed~\cite{batchNormOriginal} that the root cause of its effectiveness is a reduction of ICS.
However, experiments conducted in~\cite{explainBN} show largely unchanged behavior of networks with BN is when non-zero mean and non-unit variance \textit{i.i.d noise} (\textit{mini-batches} are assumed to be \textit{i.i.d} for BN to work) is added to their intermediate activations.
If removed ICS was the main factor in the effectiveness of BN, then explicitly adding ICS back should disturb the training process.
ICS matters, since training a neural network is a minimization problem, where the objective is to minimize the difference between the actual and desired output of the network.
Since learning is typically organized into batches of data, problems arise if the distribution of data points in subsequent batches differs too much.
Hence, what is learned from one batch can be completely unsuitable for the rest of data.

Zero-centering is helpful to find the natural distribution of data, and variance is required to keep the network from over-learning a specific set of data points.
As mentioned previously, \cite{explainBN} discusses that although the connection between reduction of ICS and performance of BN is not direct, the normalization of distributions might have a positive effect on the optimization landscape.
The authors further explore other reasons for the exceptional performance of BN, one of them is related to smoothing the optimization path.
Reparametrization, which takes place because of BN layers, results in gradients with smaller magnitude (when compared with non-BN networks).
Besides, it also smooths the loss gradients and makes them more \textit{Lipschitz}\footnote{A Lipschitz function can be defined as $|f(x)-f(y)| \leq M |x-y|$.
The slope of the secant line joining $(x,f(x))$ and $(y,f(y))$ has an upper bound represented by constant $M$.
This condition is important as the slope of the secant defines the rate of change of a function and should not grow out of bounds.},
and this indeed proves to be very helpful for gradient-based training methods.
The increased \textit{Lipschitzness} of the gradients further help in using larger learning rates and in avoiding the problem of vanishing and exploding gradients.
Decreasing the unpredictability of the gradient movement, BN contributes in maintaining the trajectory for gradient descent.

Experiments conducted in \cite{explainBN} establish that equating the performance of BN and its role in reducing ICS is incorrect, as BN might not even be contributing towards the reduction of shift in activations.
As discussed earlier, the nature of BN to reparameterize the gradients makes them well-behaved and hence enabling faster and better optimization of the network.
Another relevant work is \cite{Understanding BN}, where the authors investigate BN and its contribution towards making better networks.
It also studies the relation between large learning rates and their effect on generalization due to additional regularization of the Stochastic Gradient Descent (SGD) process.
This particular assertion stems from \cite{SGD_Discussion1}, where the authors state that the minima of loss functions calculated using SGD are not local minima or the saddle points of the loss function in the traditional sense, but are mainly governed by the learning rate and mini-batch size.
The deviation from the actual minima is controlled by the ratio $\frac{\eta}{\beta}$ where $\eta$ is the learning rate and $\beta$ is the mini-batch size. This value determining the divergence tends to be large in practice. One of the main authors of \cite{batchNormOriginal} addressed the issue of dependence of BN's performance on mini-batch size in \cite{renormalize_BN}, which uses the calculation of additional affine transforms, which act as corrective measures for the mean and variance calculated over a mini-batch.

The work further discusses that SGD does not directly operate on the loss function, but instead on a potential function which might or might not be the same as the original loss function.
This potential function is governed by the learning rate $\eta$ and mini-batch $\beta$, making it more dependent on the dataset and the architecture in consideration.
The regularization capabilities of the SGD are also determined by $\frac{\eta}{2\beta}$ and if this ratio starts approaching zero the generalization also starts to deteriorate. Hence, using a large learning rate or using a small mini-batch would have the same effect on the generalization capabilities of SGD whereas, BN does not perform well with small mini-batches it can work well with large learning rates which leads to better results in networks which use gradient based methods such as SGD.
%\todo{corollary: is a larger value is beneficial for generalization?}
This again leads to a hypothesis that the effectiveness of BN may be explained independent of ICS.
%\todo[inline]{Can either increase $\eta$ or decrease $\beta$ to improve generalization. BN helps with high $\eta$, but doesn't like small $\beta$?}

The following sections detail our experiments on training networks with and without BN, and the methods required to make networks train effectively either way.

\section{Experiments and Discussion}
\subsection{Experimental Setup}
To investigate BN and non-BN methods in detail we have trained ResNet-18, ResNet-34 and ResNet-50 as proposed in \cite{resNetoriginal}, and adapted versions of all three architectures on the ImageNet dataset \cite{imageNet}.
Standard data augmentation and pre-processing steps are followed for training and validation data preparation.
Initial experiments have been conducted using the conventional monotonically decreasing learning rate schedule.
The standard number of epochs used is 90, as it has been experimentally determined that training any further does not improve the learning.
Best results have been discussed in the performance evaluation section below, additional details and the graphs for all the experiments are present in \nameref{section:appendix}.

\subsection{Network architecture} 
\label{subsection:networkArch}
\begin{figure}		
		\begin{subfigure}{.5\textwidth}
			\centering
			\includegraphics[scale=0.4]{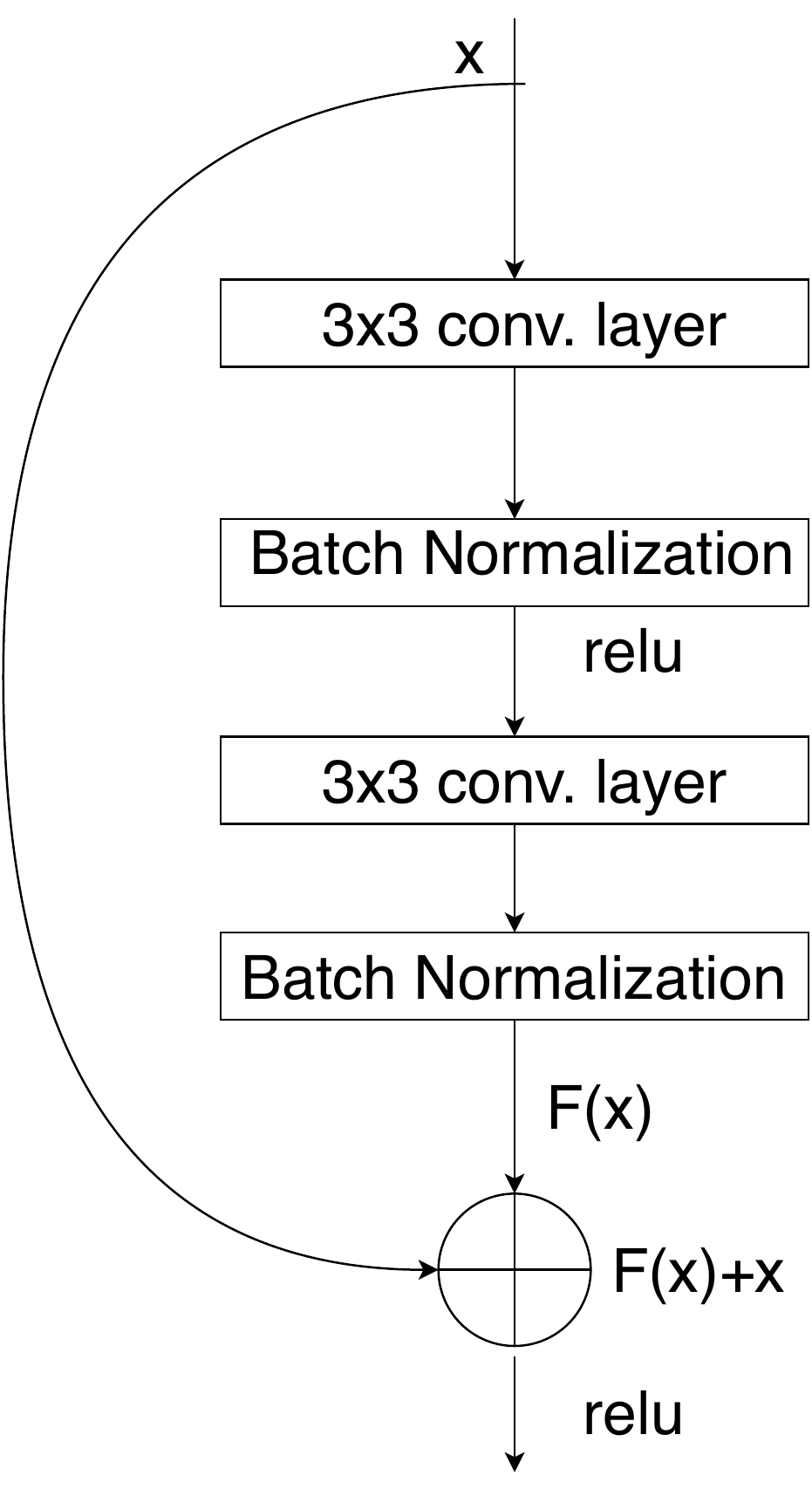}
			\caption{Original}
			\label{fig:original_ResNet}
		\end{subfigure}%
	\begin{subfigure}{.5\textwidth}
		\centering
		\includegraphics[scale=0.32]{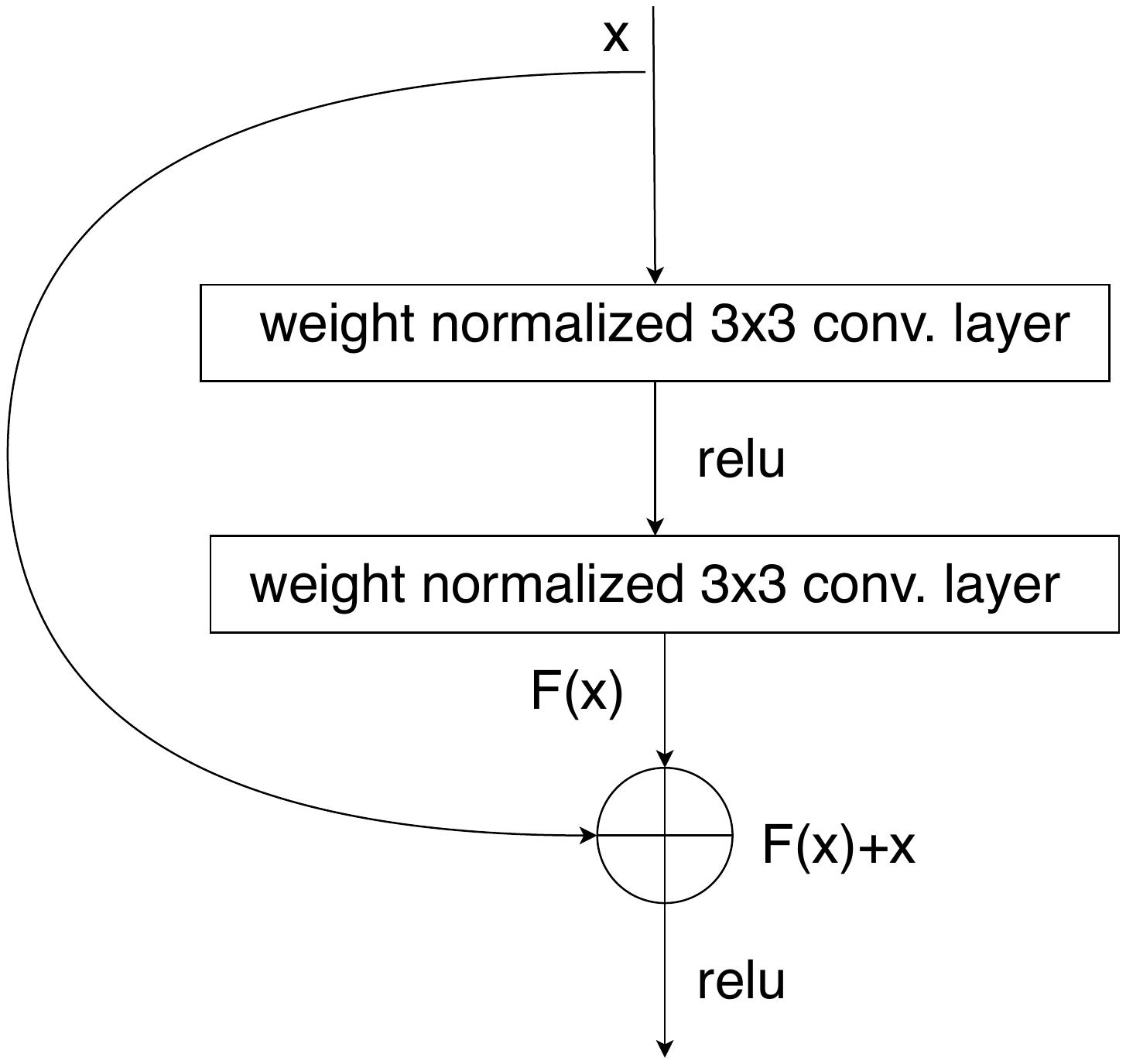}
	\caption{Modified}
	\label{fig:modified_ResNet}
	\end{subfigure}%
	\caption{Residual Learning: Basic Block}
\label{fig:ResNet-18}
\end{figure}

One of the main research questions of this work is, whether it is possible to eliminate BN layers with simpler constructs such as weight normalized convolution layers and observe the network behavior when it is subjected to large learning rates with varying learning rate schedules.
The original \textit{basic block} (\autoref{fig:original_ResNet}, used by ResNet-18 and ResNet-34) contains a combination of stacked 3x3 convolution layers, separated by a BN layer and outputs from this stack is added with the output of \textit{shortcut connection}, which performs \textit{identity} mapping in the case of residual networks.
Our proposed architecture uses the \textit{modified residual block} presented in \autoref{fig:modified_ResNet}. 
It is simpler in terms of operations and the normalization and convolutions layers are not separate. 
Consider the standard formula of a linear layer in neural network: 
\[y = \phi(w \cdot x + b)\]

Weight normalization~\cite{weightNorm} replaces the weight vector $w$ with two separate trainable parameters vector $v$ and scalar $g$. The reparameterized weight vector is represented as:

\begin{equation*}
w = {\frac{g}{\|v\|}}v
\end{equation*}

This reparametrization is done in order to separate weights into magnitude $g$ and direction $v$, as per \cite{weightNorm} this procedure contributes towards faster convergence.

Gradient clipping~\cite{gradientClipping} was developed to address the issue of exploding gradients, which proved to be a major hurdle while training recurrent neural networks.
The idea behind considering gradient clipping as one of the possible methods of normalization is to address the first question
\begin{displayquote}
Is it possible to achieve similar performance without BN if gradient magnitudes are maintained in a manner where they neither diminish nor grow out of bounds?
\end{displayquote}
which this work seeks to answer.
This work specifically focuses on the \textit{adaption} of the \textit{gradient threshold} so that it can evolve with the training process, potentially helping in maintaining a reasonable environment for optimization.
The second technique which will be employed in addition with weight normalization and gradient clipping is \textit{dropout} which was originally proposed in~\cite{dropoutPaper} to address overfitting in networks.
Dropout works by randomly setting values to $0$ with a given probability, preventing the network from learning excessive co-feature information.

The following section contains experimental results of using weight normalization along with adaptive gradient clipping and dropout in \textit{residual networks}.

\subsection{Performance evaluation}
In this work we have experimented with different methods of normalization and regularization such as weight normalization, adaptive gradient clipping, and dropout, respectively.
Networks implemented using these alternate methods are referred to as \textit{non-BN (non-Batch Normalization)} networks/methods.
As stated in \cite{explainBN}, using BN results in gradients with smaller magnitude as compared non-BN networks.
The section is further divided in multiple subsections starting with first versions of non-BN ResNet-18 network experiments followed by more complex implementations of non-BN on ResNet-18.
This section concludes with implementation of most successful  non-BN methods from ResNet-18 experiments on ResNet-34/50.    

\subsubsection{Effects of removing BN from a network}
In order to further understand the behavior of BN and non-BN networks it is important to study the training and validation accuracy of these networks, as it would help in analyzing the generalization and regularization capabilities of the respective networks.
\autoref{tab:Comparison Table for ResNet-18  initial} contains the top-1 accuracy for BN and initial versions of non-BN networks.
As mentioned earlier weight normalization \cite{weightNorm} serves as the basis of alternate normalization (non-BN) methods in this work, but before making it a default choice for following non-BN versions of network it was necessary to compare it with gradient clipping as mentioned in \nameref{subsection:networkArch}.
Learning rates used for BN and non-BN networks are 0.1 and 0.001 respectively and they are decreased monotonically for every epoch over the entire training phase.

Later, in the work we have used \textit{adaptive gradient clipping}, but for this set of experiments a constant value is chosen to serve as threshold.
As it can be observed in \autoref{tab:Comparison Table for ResNet-18 initial} \textit{training accuracy} of networks with constant gradient clipping threshold and weight normalization have comparable results with $\approx 2.5\%$ and $2\%$ lower accuracy than the BN network.
However, the difference in \textit{validation accuracy} for both non-BN networks is $\approx 8\%$, making it evident that BN network has much higher generalization capability than it's non-BN counterparts.
This analysis proves that BN helps a network to generalize better, which makes this an important point to be considered for developing refined non-BN versions later. 

\begin{table}
	\centering	
		\caption{top-1 accuracy for ResNet-18 with BN and initial versions of non-BN networks with weight normalization and constant gradient clipping.}
	\begin{tabular}[t]{lcc}
		\toprule[1.5pt]
		model&training&validation\\
		\midrule
		ResNet-18 & $73.01$ & $66.5$\\
		ResNet-18 + constant gradient clipping threshold & $70.50$ & $58.2$\\
		ResNet-18 + weight normalization & $71.01$ & $58.5$\\
		\bottomrule[1pt]
	\end{tabular}

	\label{tab:Comparison Table for ResNet-18 initial}
\end{table}%

\begin{figure}
	\begin{subfigure}{.5\textwidth}
		\centering
		\includegraphics[scale=0.40]{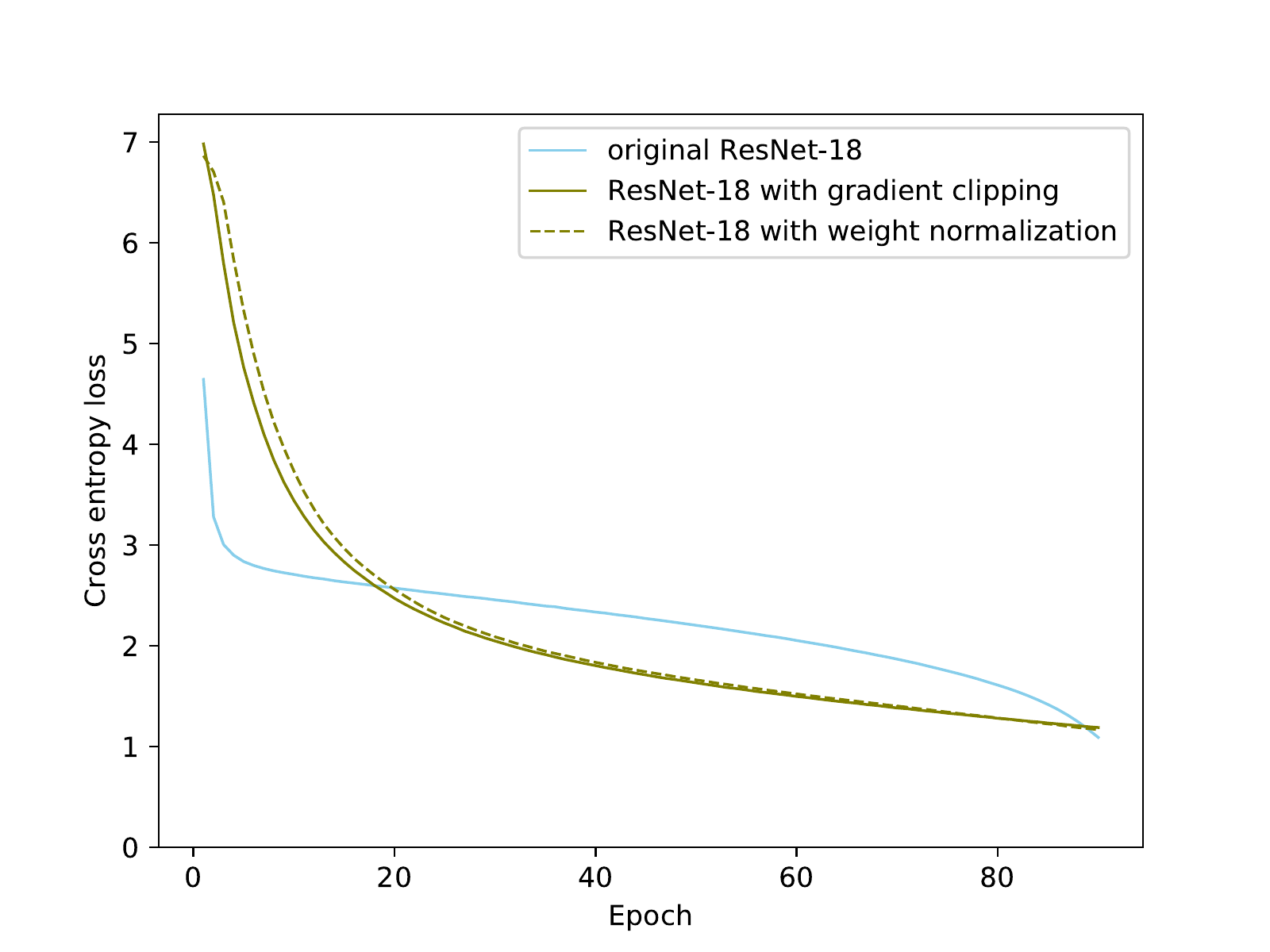}
		\caption{training phase}
		\label{fig:training loss initial}
	\end{subfigure}%
	\hfill
	\begin{subfigure}{.5\textwidth}
		\centering
		\includegraphics[scale=0.40]{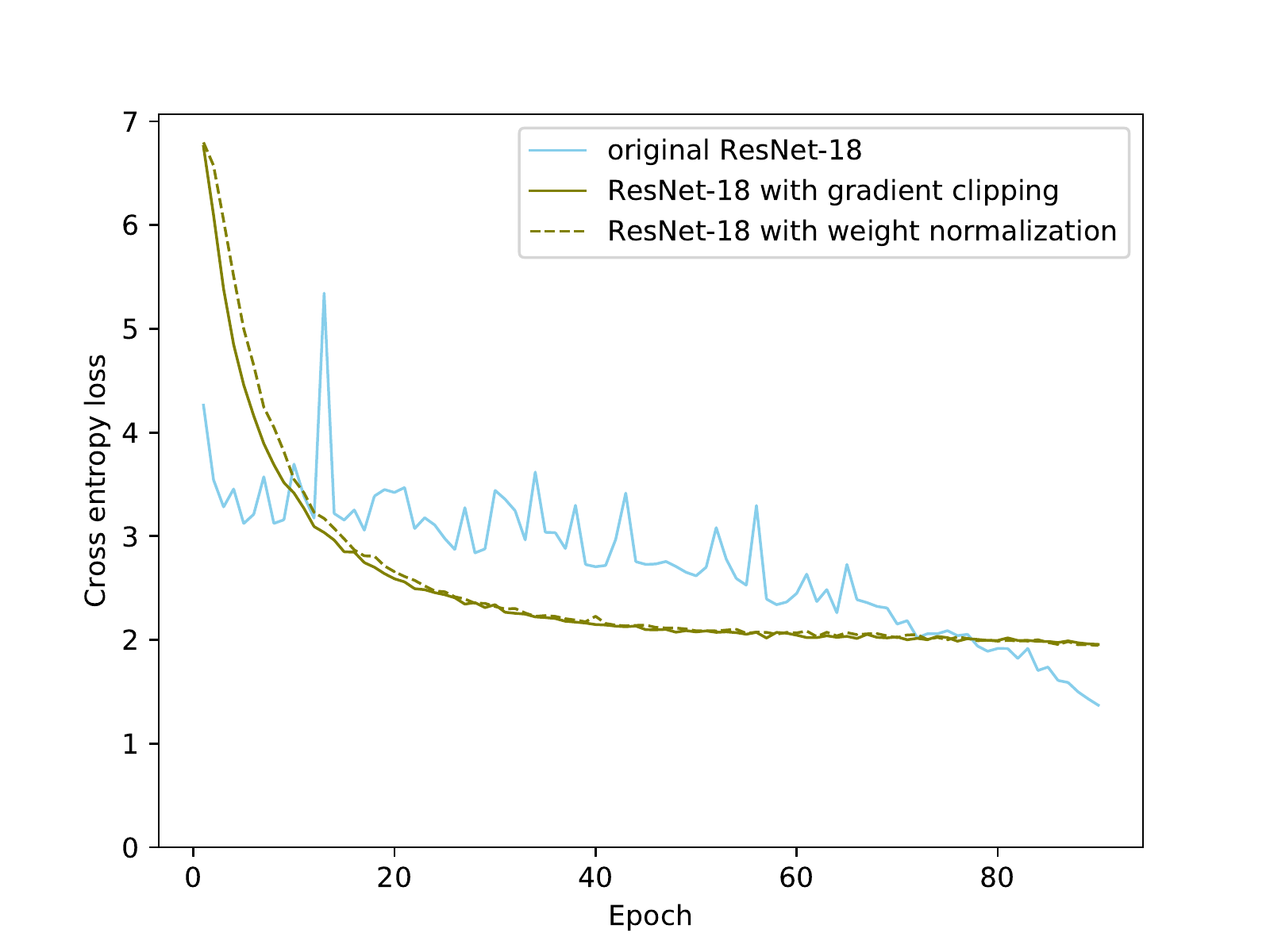}
		\caption{validation phase}
		\label{fig:validation loss initial}
	\end{subfigure}	
	\caption{Training and validation loss for BN and initial versions of non-BN models.  }
	\label{fig:figperformance_loss}
\end{figure}

\begin{figure}[h!]
	\begin{subfigure}{.5\textwidth}
		\centering
		\includegraphics[scale=0.4]{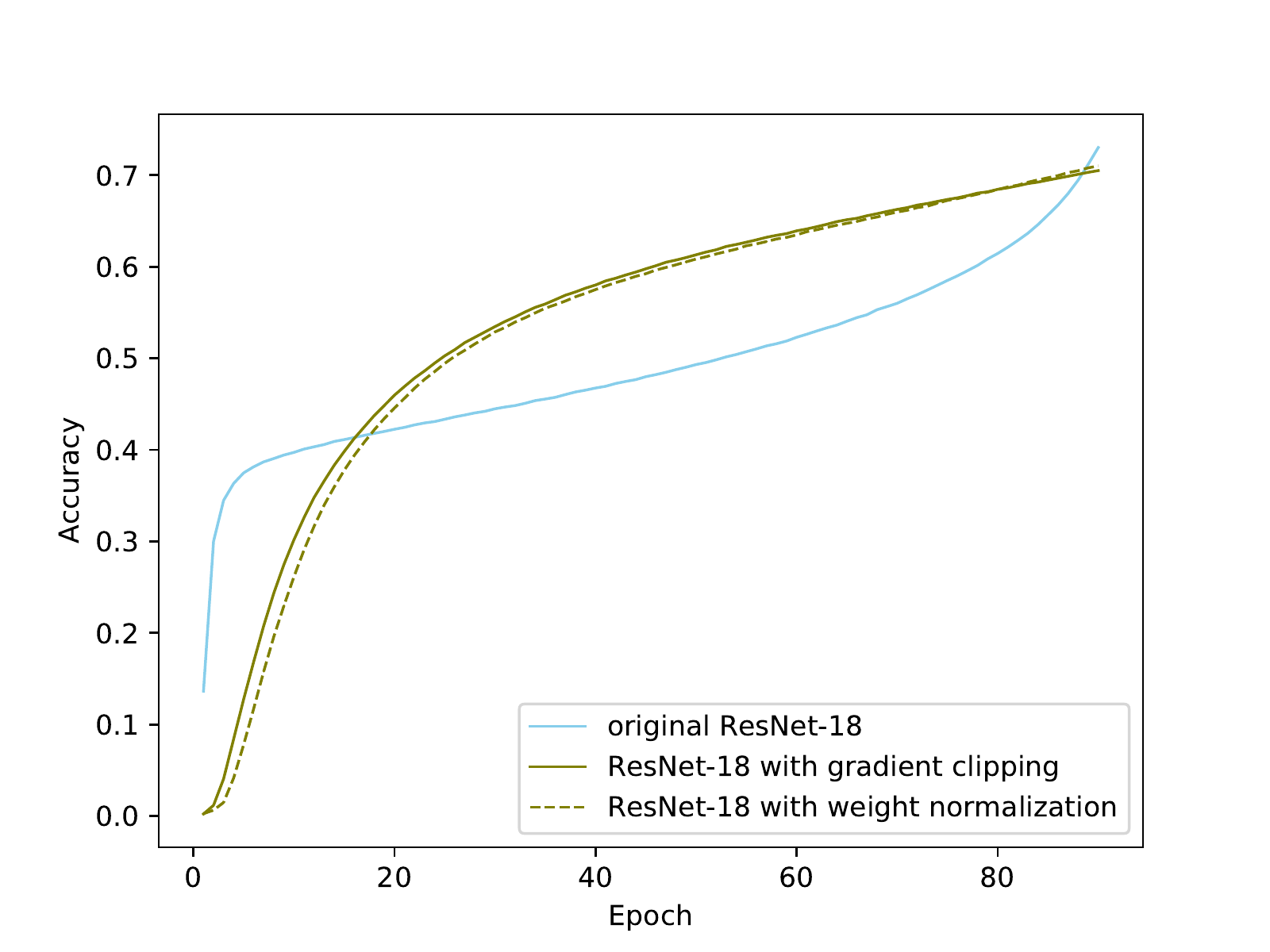}
		\caption{training phase}
		\label{fig:training_accuracy_initial}
	\end{subfigure}
	\begin{subfigure}{.5\textwidth}
		\centering
		\includegraphics[scale=0.4]{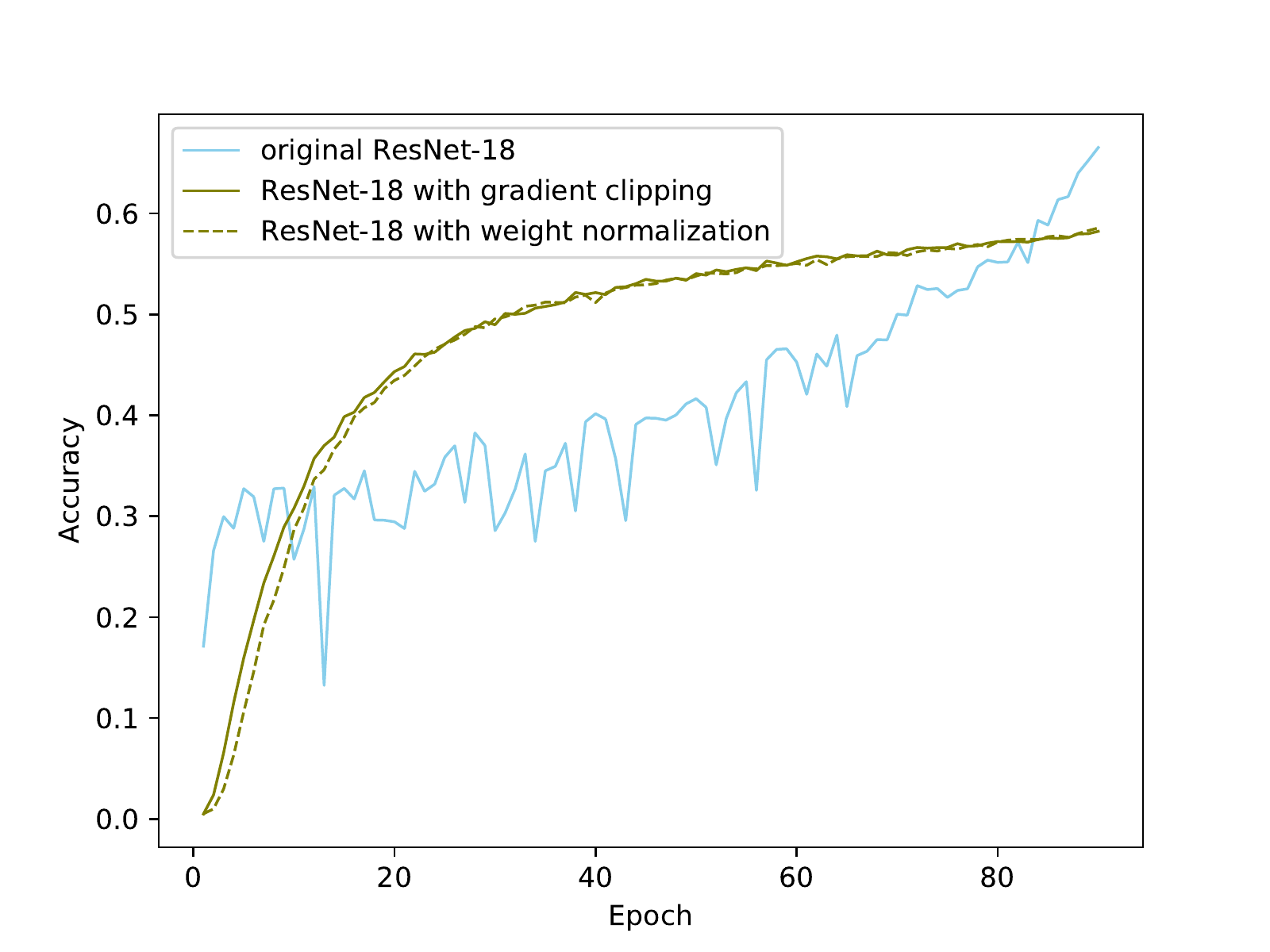}
		\caption{validation phase}
		\label{fig:validation_accuracy_initial}
	\end{subfigure}
	\caption{Comparison of training and validation phase accuracy of BN and initial versions of non-BN networks}
	\label{fig:initial}
\end{figure}

%As it is interesting to analyze the loss landscape and observe the differences between BN and initial non-BN methods.
\autoref{fig:figperformance_loss} presents the progression of the loss throughout the entire learning process.
One very prominent difference visible in both the graphs is the starting value of the loss, where the initial value for non-BN networks is higher than the BN networks.
Non-BN networks show a very steep decrease in the loss value and by the time training reached epoch 20, non-BN networks surpassed BN network.
After the rapid decrease of loss for non-BN networks in the first 20 epochs, convergence becomes slow for both BN and non-BN networks.
But from epoch 70 through 90, the BN network shows a significant improvement in validation loss, resulting in the final difference of $\approx$ 8\% between BN and non-BN networks.
The important observation which can be made from this trajectory is that the loss landscape of non-BN networks changes most significantly in the initial learning phase whereas the BN network has most significant boost in much later phases of learning.
Corresponding accuracy curves are present in \nameref{section:appendix}, \autoref{fig:initial}.

Looking at \autoref{fig:validation loss initial} it becomes clear that the initial non-BN version of the \textit{residual network} shows signs of overfitting.
Although initially validation loss looks similar to the training in \autoref{fig:training loss initial}, the non-BN networks show only a minor improvement of $\approx$ 11\% after epoch 40 until training ends after 90 epochs.
BN networks, although less stable, behave more similar to the training phase, and in the last 10 epochs validation loss decreases by $\approx$ 32\%.
These final epochs are crucial for BN networks, since they only surpass non-BN networks at this point.

\subsubsection{Improved non-BN network experiments and their comparison with corresponding BN network.}
After the analysis of results from initial experiments it has become clear that, whereas methods like weight normalization and gradient clipping with a constant threshold value work, for the training phase they still lack the generalization and regularization capabilities of the BN enabled networks. Hence, for all the other experiments in this work, non-BN networks use a combination of both weight normalization and gradient clipping.
Instead of using a constant value as threshold the value is reduced per epoch, in the logarithmic manner.
%This provides for the development of better non-BN versions of the network.
In addition to the above-mentioned changes one important measure that plays a crucial role is the choice of initial \textit{learning rate} (LR) and the method to modify it over the course of time.

LR is one of the hyper-parameter which influences the parameter update in gradient-based learning. The parameter update rule for stochastic gradient descent can be described as follows:
\begin{equation*}
p_{k} = p_{k-1} - \eta_{k} \nabla_p f(x_{k}; \theta)
\end{equation*}

The goal is to minimize the value of function $f$ (a cost function) parametrized by $\theta$ for input $x_k$ in the $k$\textsuperscript{th} iteration.
$p_k$ denotes the state of parameter $p \in \theta$ in iteration $k$.
$\eta_k$ is the LR and $\nabla_p$ denotes the gradient of $p$ for with respect to $f$.

\begin{figure}
	\centering
	\includegraphics[scale=0.4]{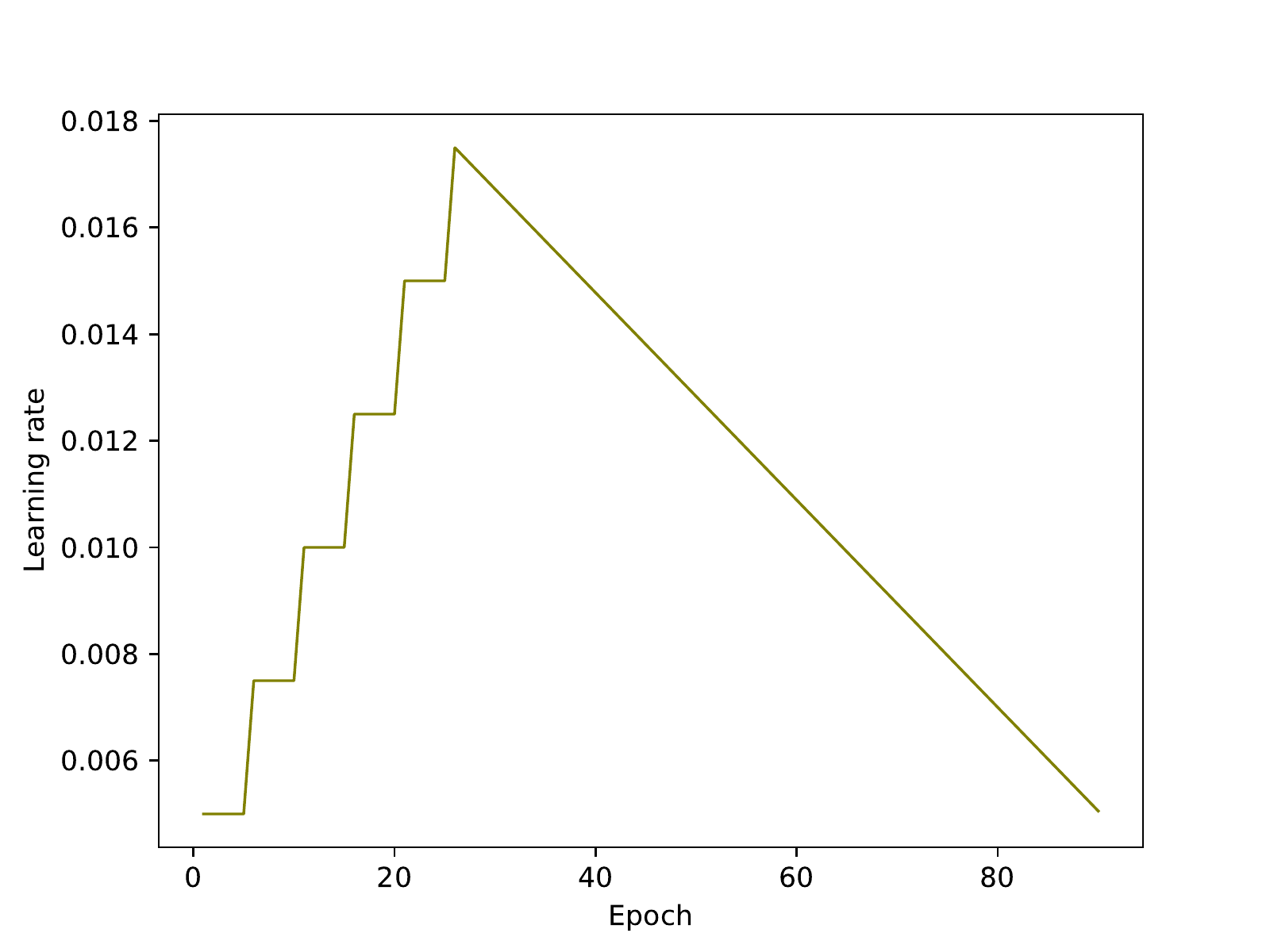}
	\caption{warm-up learning schedule}
	\label{fig:warm-up_schedule}
\end{figure}

The choice of initial LR greatly influences the optimization process.
It can be deduced that very low LR would result in slow convergence, whereas extremely high rate could cause divergence from the optimization path \cite{tuningHyperParameters}.
Hence, in addition to using weight normalization, adaptive gradient clipping and dropout this work also experiments with various other LR schedules such as cyclic schedule as described in \cite{cyclicalLR}, warm-up \footnote{This learning schedule is to start with a lower learning rate than the desired rate and stepping up until that value is reached. Followed by the standard monotonically decreasing rate. } schedule from \cite{warmUpSchedule}, which looks like \autoref{fig:warm-up_schedule} during experiments.
In addition to trying different LR schedules the initial value of the LR is also varied and it's effects are analyzed on the non-BN networks.

\begin{table}[t]
	\centering
	\caption{top-1 training and validation accuracy of BN and refined versions of non-BN networks(ResNet-18). All the networks are trained using the standard decreasing learning rate schedule. The table contains data for vanilla network which uses Batch Normalization and non-Batch Normalization versions using weight normalization (non-BN) with adaptive clipping with or without dropout layers.}
	\begin{tabular}[t]{lcccc}		
		\toprule[1.5pt]
		model&training&validation&LR\\
		\midrule
		batch normalized &73.01&66.5&0.1\\
		\midrule
		non-BN with adaptive clipping \& dropout&72.34&64.7&0.01\\
		non-BN with adaptive clipping&79.3&63.6&0.01\\
		non-BN with adaptive clipping &76.25&59.6&0.001\\
	
		\bottomrule[1pt]
	\end{tabular}
	
	\label{tab:Comparison Table for appendix(ResNet-18)}
\end{table}

\begin{table}[b]
	\centering
	\caption{Accuracy vs initial clipping value for ResNet-18}
	\begin{tabular}[t]{lccc}
		\toprule[1.5pt]
		Initial clip value&training&validation\\
		\midrule
		5 &72.3&64.7\\
		\midrule
		10&72.2&64.6\\
		\bottomrule[1pt]
	\end{tabular}
	
	\label{tab:logClip_graph}
\end{table}

\autoref{tab:Comparison Table for appendix(ResNet-18)} shows the accuracy of BN and non-BN networks using same LR schedule but different initial \textit{learning rate}.
\textit{Gradient clipping} is addressed as \textit{adaptive} because \textit{threshold} has been modified per epoch.
For the experiments presented in this work, gradient \textit{threshold} has been increased in a logarithmic manner.
In prior case study a range search was conducted to deciding the starting value of the \textit{gradient clipping threshold}.
After experimenting with values 1, 5 and 10, the convergence of the network with clipping value 1 was too slow to train properly and 5 as the initial value performs slightly better than 10 (see \autoref{tab:logClip_graph}), hence it has been used as a threshold for all the following experiments. 
%\todo{did threshold=1 not converge? if so, mention it in the table as well}%
This leads to another important observation that even though two networks were trained with different initial clipping threshold, the final results were almost the same.
Hence, if initial values belong to a certain range then the final accuracy can be similar.

Further, results in \autoref{tab:Comparison Table for appendix(ResNet-18)} show that network with dropout layers in addition to the weight normalization and adaptive gradient clipping converged with the so far highest learning rate for non-BN network of $0.01$ and yields better results when compared with ones trained at without dropout and the same or lower \textit{learning rate}.  Related graphs are present in \nameref{section:appendix}, \autoref{fig:with_without_dropout}, the section also contains a brief discussion and the motivation behind the choice of one particular way of implementing the non-BN networks.

\autoref{tab:Comparison Table for cyclic(ResNet-18)} contains results obtained after using cyclic LR schedule from \cite{cyclicalLR}. The core concept of cyclic LR is to start from a smaller \textit{learning rate} which acts as the minimum value, monotonically updating the rate and reaching the maximum desired value.
Once the maximum value is reached the \textit{learning rate} is set to decay until it again reaches the minimum.
This forms a \textit{triangular cyclic schedule} for updating learning rate.
It can be observed that the final accuracy of network worsens when using a higher learning rate while using cyclic LR schedule.
The step decrease \footnote{constant learning rate for given number of initial epochs, and standard decay afterwards.} gives better results than the cyclic learning schedule, but the accuracy of non-BN networks trained with the previous schedule as shown in \autoref{tab:Comparison Table for appendix(ResNet-18)} is several percentage points higher.
This rules out the possibility of using these two learning schedules for non-BN networks. 

\begin{table}[t]
\centering
		\caption{top-1 training and validation accuracy of BN and refined versions of non-BN networks (ResNet-18). The table contains data for vanilla network which uses Batch Normalization (BN) and non-Batch Normalization versions using weight normalization (non-BN) with adaptive gradient clipping and dropout layers.}
		
	\begin{tabular}[t]{lcccc}
		\toprule[1.5pt]
		model&LR schedule&LR&training&validation\\
		\midrule[1pt]
		 batch normalized&monotonic decrease&0.1&73.01&66.5\\
		\midrule
		non-BN with adaptive clipping \& dropout&monotonic decrease&0.01&72.34&64.7\\
		non-BN with adaptive clipping \& dropout &cyclic&0.01&54.87&54.87\\
	    non-BN with adaptive clipping \& dropout &cyclic&0.001&64.65&59.7\\
		non-BN with adaptive clipping \& dropout &step decrease&0.001&68.9&61.95\\	
		 	
		\bottomrule[1pt]
	\end{tabular}

	\label{tab:Comparison Table for cyclic(ResNet-18)}
\end{table}

As observed from the previous experiments, there are certain training methods which work for non-BN networks and some which fail to aid the learning process of networks without BN.
The highest learning rate which could be reached with non-BN networks was \textit{0.01}, whereas for BN network it can be as high as \textit{0.1}.
It is important to figure out the highest learning rate which can be supported by non-BN networks before they start diverging to improve generalization~\cite{SGD_Discussion1} as mentioned earlier.

\begin{figure}[t]
	\begin{subfigure}{.5\textwidth}
		\centering
		\vspace*{-2em}\includegraphics[scale=0.4]{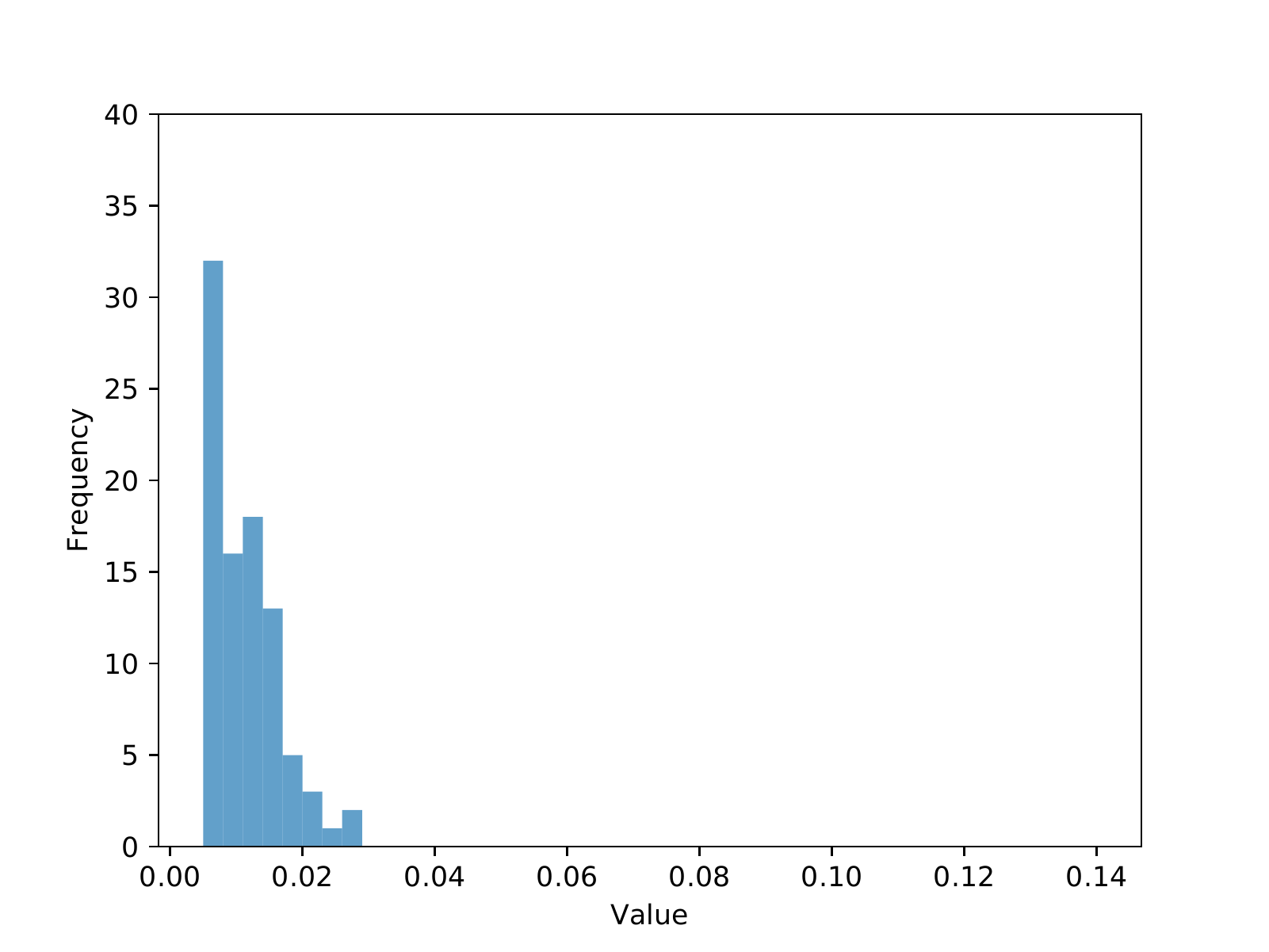}\vspace*{-0.5em}
		\caption{ResNet 18 trained with no normalization and regularization method}\vspace*{-0.2em}
		\label{fig:unNormalized Network}
	\end{subfigure}
	\begin{subfigure}{.5\textwidth}
		\centering
		\vspace*{-2em}\includegraphics[scale=0.4]{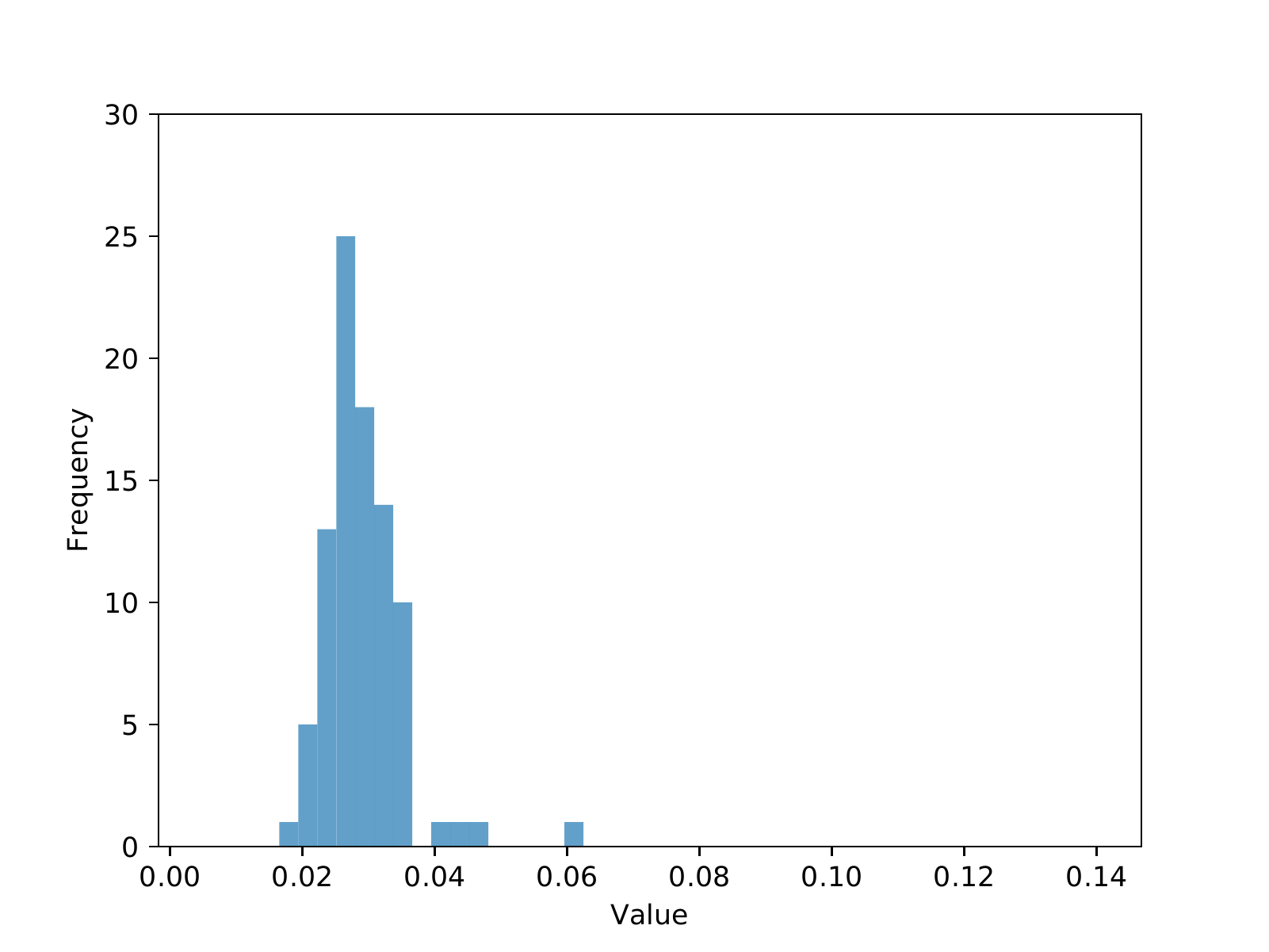}\vspace*{-0.5em}
		\caption{Vanilla ResNet 18 with BN and LR 0.1}\vspace*{1.0em}
		\label{fig:Vanilla resnet 18 with BN and LR 0.1}
	\end{subfigure}%
	\hfill
	\begin{subfigure}{.5\textwidth}
		\centering
		\vspace*{-1em}\includegraphics[scale=0.4]{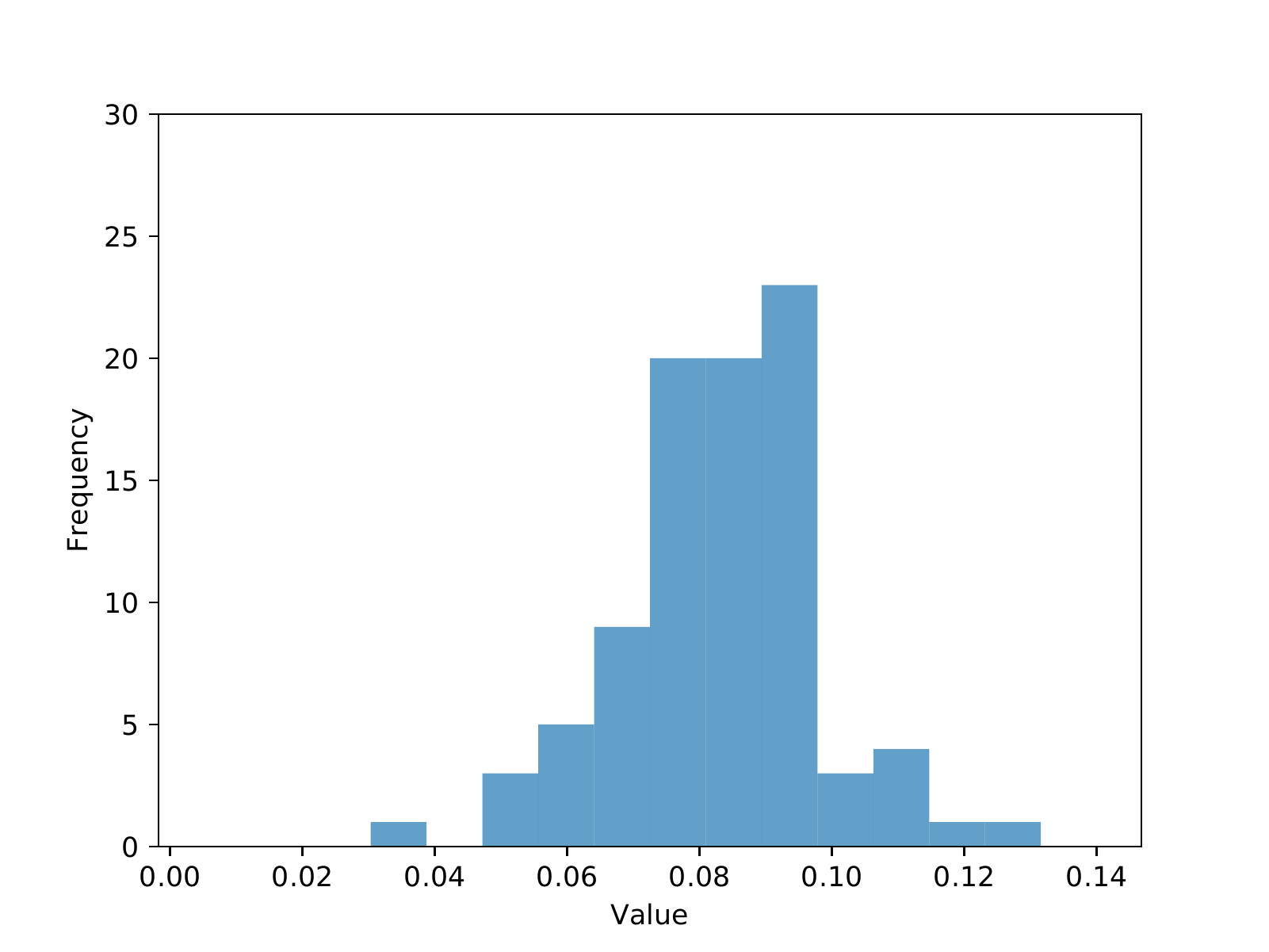}\vspace*{-0.5em}
		\caption{ResNet 18 trained with weight normalization, gradient clipping and LR 0.01}
		\label{fig:weightnorm and norm clipping with LR 0.01}
	\end{subfigure}
	\begin{subfigure}{.5\textwidth}
		\centering
		\vspace*{-1em}\includegraphics[scale=0.4]{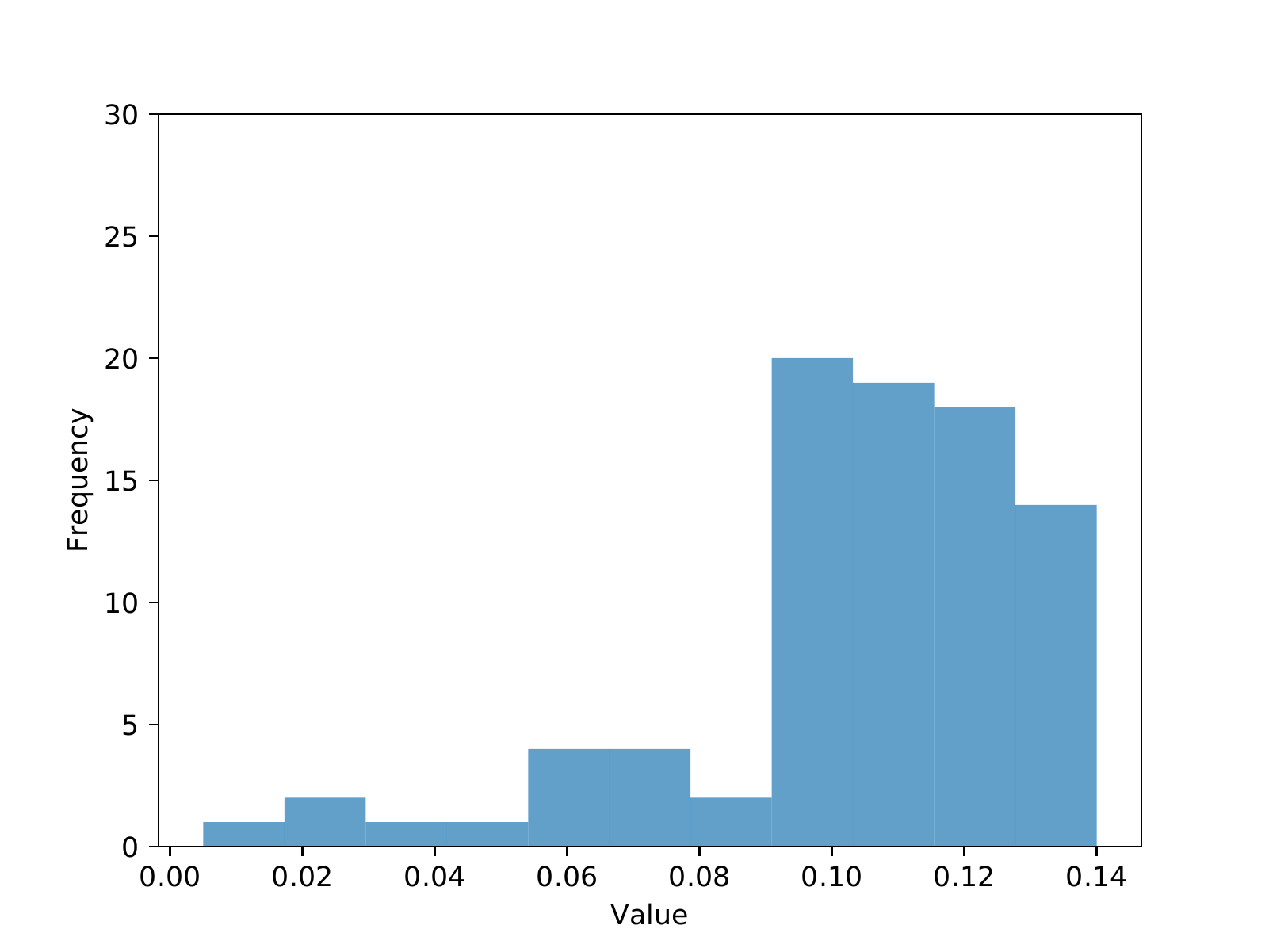}\vspace*{-0.5em}
		\caption{ResNet 18 trained with weight normalization, gradient clipping and LR 0.001}
		\label{fig:weightnorm and norm clipping with polyPolicy LR 0.001}
	\end{subfigure}
	
	\caption{Gradient distribution for ninth layer of ResNet-18, for un-normalized, BN and non-BN networks. These observations are made using monotonically decreasing learning rate.}
	\label{fig:fig}
\vspace*{-1em}
\end{figure}

\subsubsection{Gradient analysis for BN and non-BN networks}
To better understand how training differs with and without BN, as well as other measures to improve training, the magnitudes of gradients were logged.
\autoref{fig:fig} shows histograms of the distribution of gradient magnitudes for the ninth layer in ResNet-18 during the entire training phase.
\autoref{fig:unNormalized Network} represents an unnormalized network trained with the highest possible learning rate $0.00001$.
With these settings the network neither diverged, but did also not converge to a meaningful state, resulting in a network which has a performance similar to random selection. Observed magnitudes were mostly very low, hinting towards a vanishing gradient problem.
Hence, it can be asserted that without BN even \textit{residual networks} need additional normalization and regularization methods.
\autoref{fig:Vanilla resnet 18 with BN and LR 0.1} represents the distribution for a network trained with BN at learning rate $0.1$.
Finally, \autoref{fig:weightnorm and norm clipping with LR 0.01} and \autoref{fig:weightnorm and norm clipping with polyPolicy LR 0.001} represent the gradient distributions for the modified network, with learning rate of $0.01$ and $0.001$ respectively.
Ignoring the unnormalized network which did not converge, it can be observed that the network with BN has gradients with smaller magnitude, when compared with modified network even when the learning rate is kept relatively higher for BN based networks.
The second observation is the distribution of magnitude of gradients for the BN network is roughly centered around the mean value, whereas the non-BN networks are skewed towards higher values and much greater standard deviation.

\begin{table}[t]
	\centering
	 \caption{top-1 accuracy of non-BN network trained with adaptive momentum and warm-up LR schedule from \cite{warmUpSchedule}}
	\begin{tabular}[t]{lcc}
		\toprule[1.5pt]
		model&training&validation\\
		\midrule
		batch normalized&73.01&66.5\\	
		non-BN with adaptive clipping \& dropout&74.7&65.1\\
		\bottomrule[1pt]
	\end{tabular}
   
	\label{tab:Comparison Table for ResNet-18}
\end{table}

\subsubsection{Refined non-BN ResNet-18 network}
A final set of experiments was curated for ResNet-18 architecture, some changes which help \textit{residual networks with BN} from \cite{warmUpSchedule} were implemented for \textit{non-BN residual networks}. The important part was the use of warm-up \textit{learning rate} schedule and see if the effect of BN on the change of loss in last 10 epochs as shown in \autoref{fig:figperformance_loss} can be produced for non-BN networks. Updating the momentum along with the learning rate was also implemented but no improvement was observed. \autoref{tab:Comparison Table for ResNet-18} shows the results from BN network and non-BN network.
This is the best result achieved by any non-BN networks discussed above.

\begin{figure}[b!]
\vspace*{-2em}
	\begin{subfigure}{.5\textwidth}
		\centering
		\includegraphics[scale=0.40]{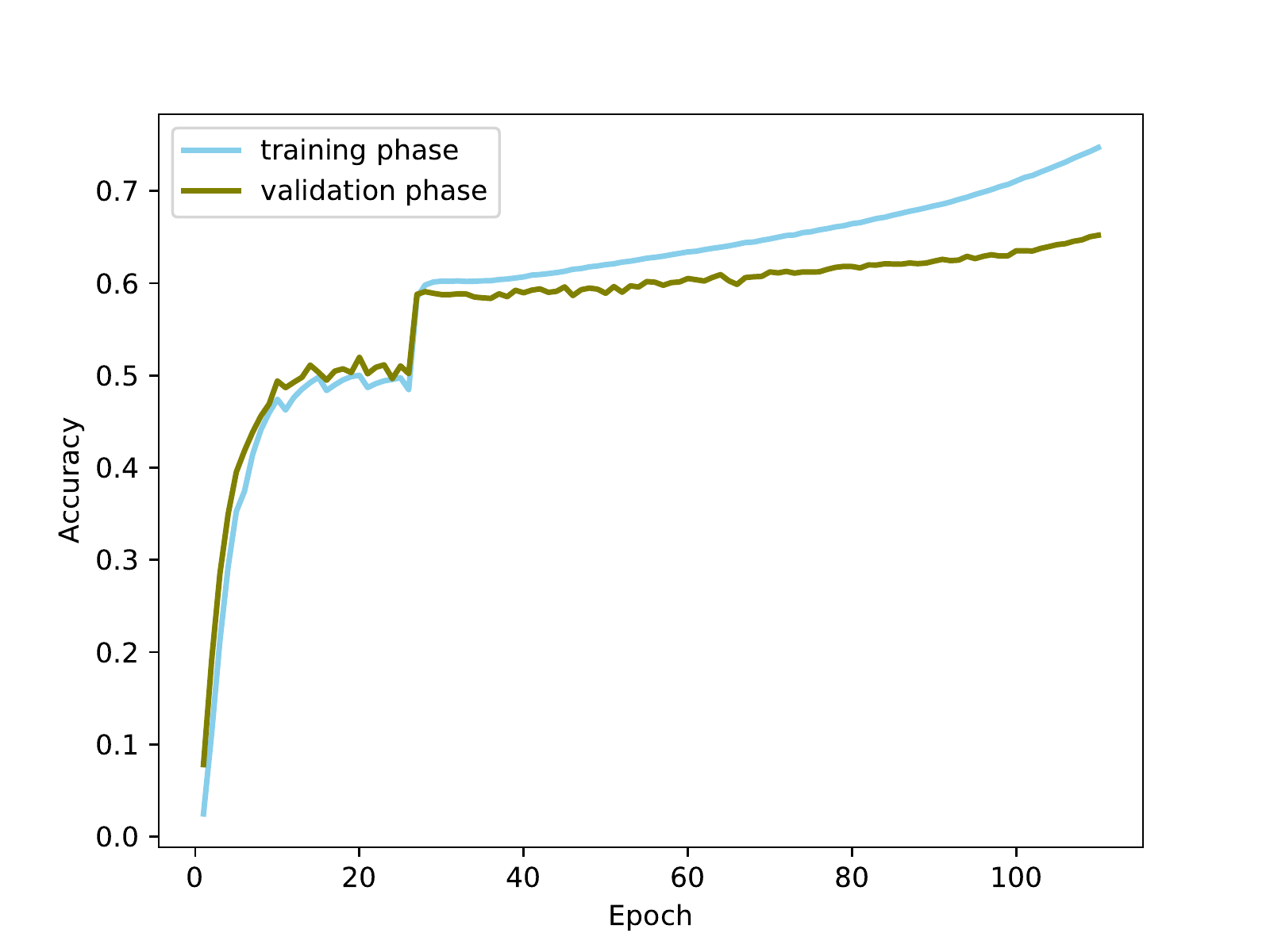}
		\caption{combined accuracy}
		\label{fig:combined accuracy}
	\end{subfigure}%
	\hfill
	\begin{subfigure}{.5\textwidth}
		\centering
		\includegraphics[scale = 0.40]{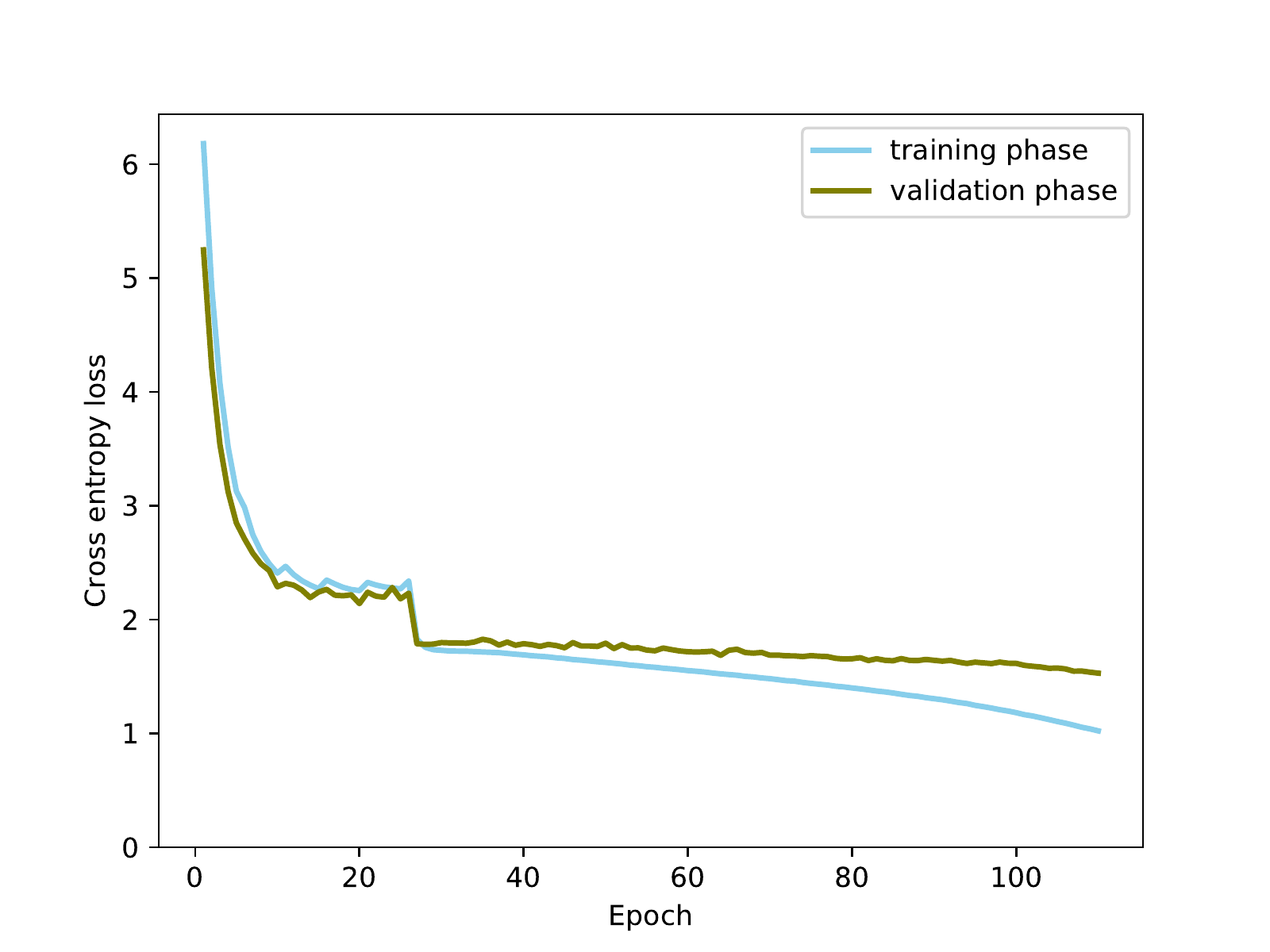}
		\caption{combined loss}
		\label{fig:combined loss}
	\end{subfigure}	
\vspace*{-0.5em}
	\caption{top-1 loss and accuracy of non-BN network with warm-up LR and momentum correction from \cite{warmUpSchedule} }
	\label{fig:figperformance_warmUp}
\end{figure}

The non-BN network mentioned was trained for 120 epochs as opposed to 90 epochs used for all the previous experiments. Motive behind this choice was to give enough time for the learning rate to decay completely. The graphs in \autoref{fig:figperformance_warmUp} show the progression of loss and accuracy in during learning process. It can be seen clearly that there is a sudden change in the accuracy and loss values around epoch 25, this occurs due to change in learning rate at that point. The highest \textit{learning rate} this network can achieve is \textit{0.017}.

\subsubsection{Performance analysis of BN and refined non-BN ResNet-34/50 networks}
While ResNet-18 is convenient to study due to its relatively low complexity compared to other architectures, it is also relatively shallow by modern standards, thus making it necessary to test whether the presented techniques generalize to deeper residual networks such as ResNet-34 and 50.
The experiments and methods which bore best results from the ResNet-18 architecture are implemented on these deeper networks.
\autoref{tab:Comparison Table for ResNet-50} summarizes the results from selected experiments.
The first difference which can be observed is that, unlike ResNet-18, the dropout layer's presence does not contribute much to the end results.
Non-BN variants for ResNet-34 and 50 exhibited more over-fitting compared to the shallower network.
%the training accuracy for all networks is $\approx$ 10\% more and validation accuracy is $\approx$ 10\% less than the original ResNet-50.
While ResNet-34 at least behaved broadly similar to previous experiments, achieving respectable accuracies, the same cannot be said about ResNet-50
Even the changes including adaptive momentum and warm-up LR schedule which worked for ResNet-18 appear to not be beneficial for training ResNet-50, although the non-BN versions did not fail completely and might be improved further from hereon out.	

\begin{table}[t]
	\centering
		\caption{top-1 accuracy for ResNet-18/34/50. \textit{Learning rate} used for all the non-BN networks are 0.01 for monotonically decreasing \& 0.005 for warm-up schedule. The table contains data for vanilla network which uses Batch Normalization and non-Batch Normalization versions using weight normalization (non-BN) with adaptive gradient clipping and dropout layers.}
	\begin{tabular}[t]{lccc}
		\toprule[1.5pt]
		model&training&validation&LR schedule\\
		\midrule
		ResNet-18\\
		\midrule
		batch normalized&73.01&66.5&monotonically decreasing\\
		\midrule
		non-BN with adaptive clipping \& dropout&72.34&64.7&monotonically decreasing\\
		&74.7&65.1&warm-up\\	
		\midrule[1.0pt]
		ResNet-34\\
		\midrule
		batch normalized&77.6&70.1&monotonically decreasing\\
		\midrule
		non-BN with adaptive clipping \& dropout&75.5&67.4&monotonically decreasing\\
		&75.2&68.01&warm-up\\
		\midrule[1.0pt]
		ResNet-50\\
		\midrule
		batch normalized&79.4&72.6&monotonically decreasing\\
		\midrule
		non-BN with adaptive clipping \& dropout&82.4&62.8&monotonically decreasing\\
		&86.01&62.22&warm-up\\
		
		\bottomrule[1.0pt]
	\end{tabular}

	\label{tab:Comparison Table for ResNet-50}
\end{table}

\subsection{Comparison of runtime and memory requirements}
All experiments, including the following measurements, were performed on Nvidia DGX-1 equipped with Tesla V100 16GB GPUs.
Pytorch version 1.4 was used to implement the networks and training process.
\autoref{tab:Comparison Table for ResNet-18 performance} shows the average time taken along with the peak memory consumption per epoch. It is evident that even though the peak memory consumption of BN network is $\approx$ 5 times higher, time taken per epoch is around 1.5 times less.
Further investigation shows that weight normalization of the convolution layers is the biggest contribution factor to longer epoch times.
This is counter-intuitive at first, since weight normalization is a relatively lightweight operation on the convolution weights, which are much lower volume compared to the intermediate tensors that are normalized by BN.
However, since BN is so prevalent much development time has been spent on optimizing it.
Pytorch uses functions provided by cuDNN for best performance on the used hadware.
The same cannot be said about weight normalization, which is implemented through standard tensor math functions.
At this point in time there exists a trade-off between memory consumption and time taken for training when using non-BN methods.

\begin{table}[t]
	\centering	
		\caption{time and peak GPU consumption per epoch for various normalization methods used with ResNet-18. The table contains data for vanilla network which uses Batch Normalization and non-Batch Normalization versions using weight normalization (non-BN) with or without other adaptations.}
	\begin{tabular}[t]{lcc}
		\toprule[1.5pt]
		model&epoch (minutes)&memory (GB)\\
		\midrule
		vanilla &16&7\\
		non-BN &21&1.4\\
		non-BN with adaptive clipping&23&1.4\\
		non-BN with adaptive clipping \& dropout&26&1.5\\	    		
		\bottomrule[1pt]
	\end{tabular}

	\label{tab:Comparison Table for ResNet-18 performance}
\end{table}%
 
\section{Conclusion} 
This work was undertaken to address three important questions related to the use of \textit{batch normalization} (BN) in neural networks. 
The first two questions this work attempts to answer are:

\textit{Is it possible to achieve BN like performance if the gradient magnitudes are maintained in a manner where they neither diminish nor grow out of bounds?}
and
\textit{Is it possible to achieve the same network performance, by replacing BN layers with simpler constructs such as weight normalized convolution layers, adaptive gradient clipping and dropout?}

\autoref{fig:fig}, contains gradient distributions for four versions of ResNet-18 including unnormalized, vanilla (using BN), non-BN with LR 0.01, and non-BN with LR 0.001.
Without normalization gradients are greatly diminished, which potentially contributed towards the network not converging.
Gradients of vanilla network are more normally distributed and have significantly higher magnitudes when compared with the unnormalized version, with low standard deviation and high symmetry.
Whereas, for both non-BN yet normalized network versions,  the distribution mean is much higher and it is further heavily skewed towards higher values.
However, relatively high accuracies achieved by these networks shows that gradients were controlled enough to maintain a reasonable optimization landscape for non-BN networks.

To address the second question experiments were performed for vanilla ResNet-18 and ResNet-18 with \textit{modified basic block}.  \autoref{tab:Comparison Table for ResNet-18} shows the results of these experiments.
The non-BN network presented in this table got best results with performance of validation phase for modified architecture $\approx$ 1.4\% lower than the original residual network.

Apart from modifying network architecture, various other adaptations were considered for training networks. 
One of the most important choices was the \textit{learning rate schedule}.
At the beginning of this work it was assumed that because of \textit{cyclic learning rate} schedule's nature it would be most suitable for non-BN networks.
But on the contrary it proved to be unsuitable, which can be observed in \autoref{tab:Comparison Table for cyclic(ResNet-18)}.
Similarly to its contribution to large batch training, a learning rate schedule with a warm-up phase and subsequent linear decay was most successful in our experiments.

After training a relatively shallow ResNet-18, further experiments were conducted to check the transferability of results to deeper network architectures such as ResNet-34 and 50.
Results from experiments conducted on ResNet-34 and ResNet-50 are present in \autoref{tab:Comparison Table for ResNet-50}.
It can be observed that whereas reasonably good results are possible for non-BN ResNet-34, also referred as \textit{modified basic block} in this work performed less by $\approx$ 2\% in both training and validation phase, the same cannot be said for ResNet-50 as the network was clearly over-fitted and generalized poorly.

While most of the non-BN networks could not achieve the results which BN can, especially for the deeper versions of \textit{residual networks} (ResNet-50) \autoref{tab:Comparison Table for ResNet-50}.
But non-BN networks trained without diverging from the optimization landscape, proving that although the modifications which are applicable to ResNet-18 and ResNet-34 might not be completely transferable to ResNet-50, a detailed analysis and modification may help training deeper residual networks better without \textit{batch normalization}.

Lastly,
\textit{do the alternate methods show the same resilience towards large learning rates, as BN does without compromising the quality of trained model?}
This poses as an important question because prior work~\cite{SGD_Discussion1} suggests that large learning rates improve network performance.
In our work non-BN networks were able to generally work with \textit{0.01} as the initial learning rate, but the network trained with warm-up \textit{learning rate schedule} was able to achieve highest learning rate of \textit{0.017}.
Even those learning rates used are very high compared to the unnormalized network, which diverged for learning rates above 0.00001 and did not converge otherwise.
It is clear that the presented non-BN networks can use relatively higher learning rates, but not high as BN networks, hence it remains an open question for future work.

Apart from the three main questions addressed above, another major evaluation criteria is the resource consumption of a neural network. 
Initially it was assumed that non-BN networks would be faster in terms of computation time due to lower computational requirements, but on the contrary time taken for these calculations was much higher when compared with equivalent BN networks.
However, peak memory consumption was only $20\%$ that of BN networks.
Results for time and GPU consumption can be observed in \autoref{tab:Comparison Table for ResNet-18 performance}.

Finally, all experiments were conducted on \textit{residual networks}, which are built and optimized around batch normalization.
While the methods presented here can be used to train residual networks without batch normalization to high accuracies, the results are still slightly lower compared to when batch normalization is used.
It is not unreasonable to assume that, given more time and effort to find better hyperparameters and network architectures, configurations could be found that further close this gap or even improve upon the baseline set by batch normalization.

% the environments 'definition', 'lemma', 'proposition', 'corollary',
% 'remark', and 'example' are defined in the LLNCS documentclass as well.

\bibliographystyle{splncs04}

\section*{Appendix A}
\label{section:appendix}
\textbf{Stochastic Gradient Descent} the concept of stochastic approximation was introduced in the work of Robbins and Monro \cite{RobbinsMonro}.
The work describes the method to find root of an increasing function f :  $\mathbb{R}$ $\to$  $\mathbb{R}$, even when only noisy estimates of values are available at any given point. if we treat f as the gradient of some function F, the Robbins-Monro algorithm can be viewed a minimisation procedure. Today the root finding and optimization methods are used in practice in the form of Stochastic Gradient Descent(SGD).\newline
The SGD, along with continuous and discrete gradient descent was elaborately discussed in the works \cite{bottou1998}, \cite{sutskever2013}. Training the neural networks is equivalent to solving a non-convex optimization problem of the following form:

\[min_{x \in R^n} f(x)\]
\begin{equation*}
\text {where } f  \text{ is the loss function.}
\end{equation*}
The term ‘stochastic’ signifies the randomness used by the algorithm to pick the samples during an iteration. The traditional gradient descent algorithm requires the entire data for calculation and hence, would prove to be a computational overhead for the datasets which are extremely large, which are typically used for the training of deep networks. Due to the manner in which the SGD is implemented the descent path taken by the algorithm is much nosier when compared with GD (gardient descent). This comes as a trade-off between the training time and staggered convergence.\newline

The benefits that SGD offers include the effective use of the information available, firstly because the data might have a lot of redundancy and using the entire data set for calculating the gradient might not be very fruitful. Secondly, as it performs faster and at less per-iteration cost in the initial stages of the training phase.

\begin{figure}
	\begin{subfigure}{.5\textwidth}
		\centering
		\includegraphics[scale=0.4]{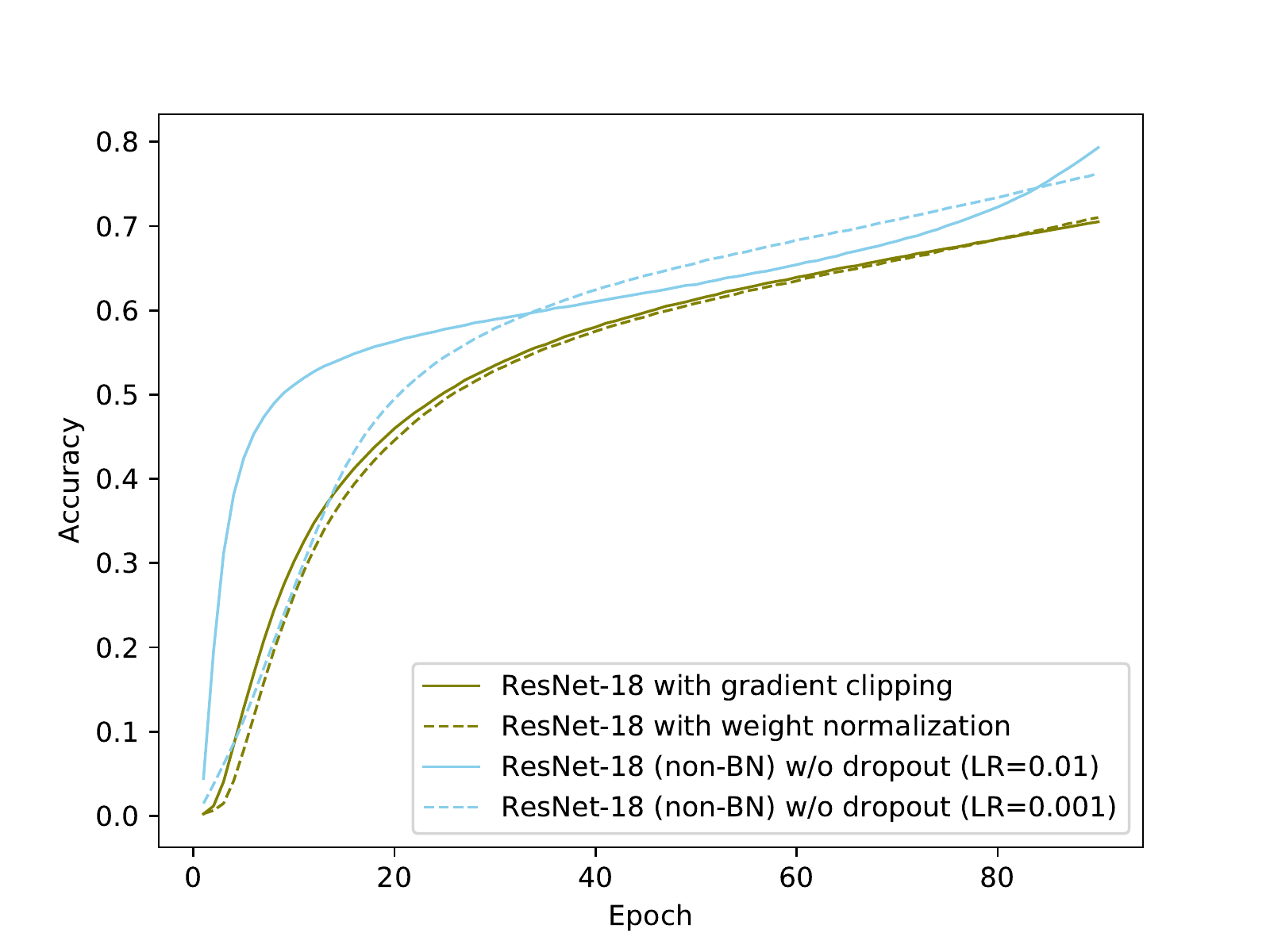}
		\caption{training accuracy}
		\label{fig:all_non_BN_training_accuracy}
	\end{subfigure}%
	\hfill
	\begin{subfigure}{.5\textwidth}
		\centering
		\includegraphics[scale=0.4]{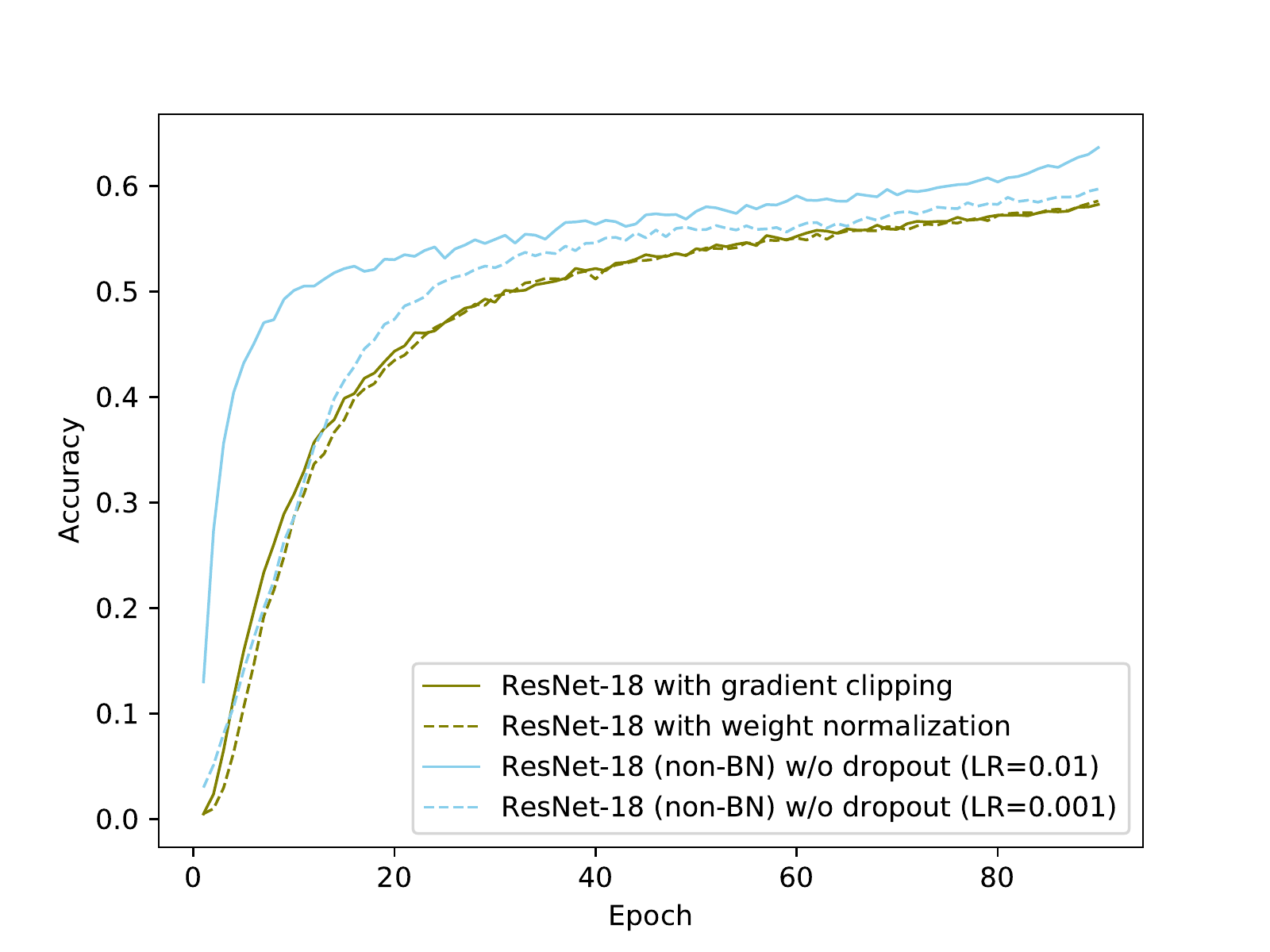}
		\caption{validation accuracy}
		\label{fig:all_non_BN_validation_accuracy}
	\end{subfigure}
	\begin{subfigure}{.5\textwidth}
		\centering
		\includegraphics[scale=0.4]{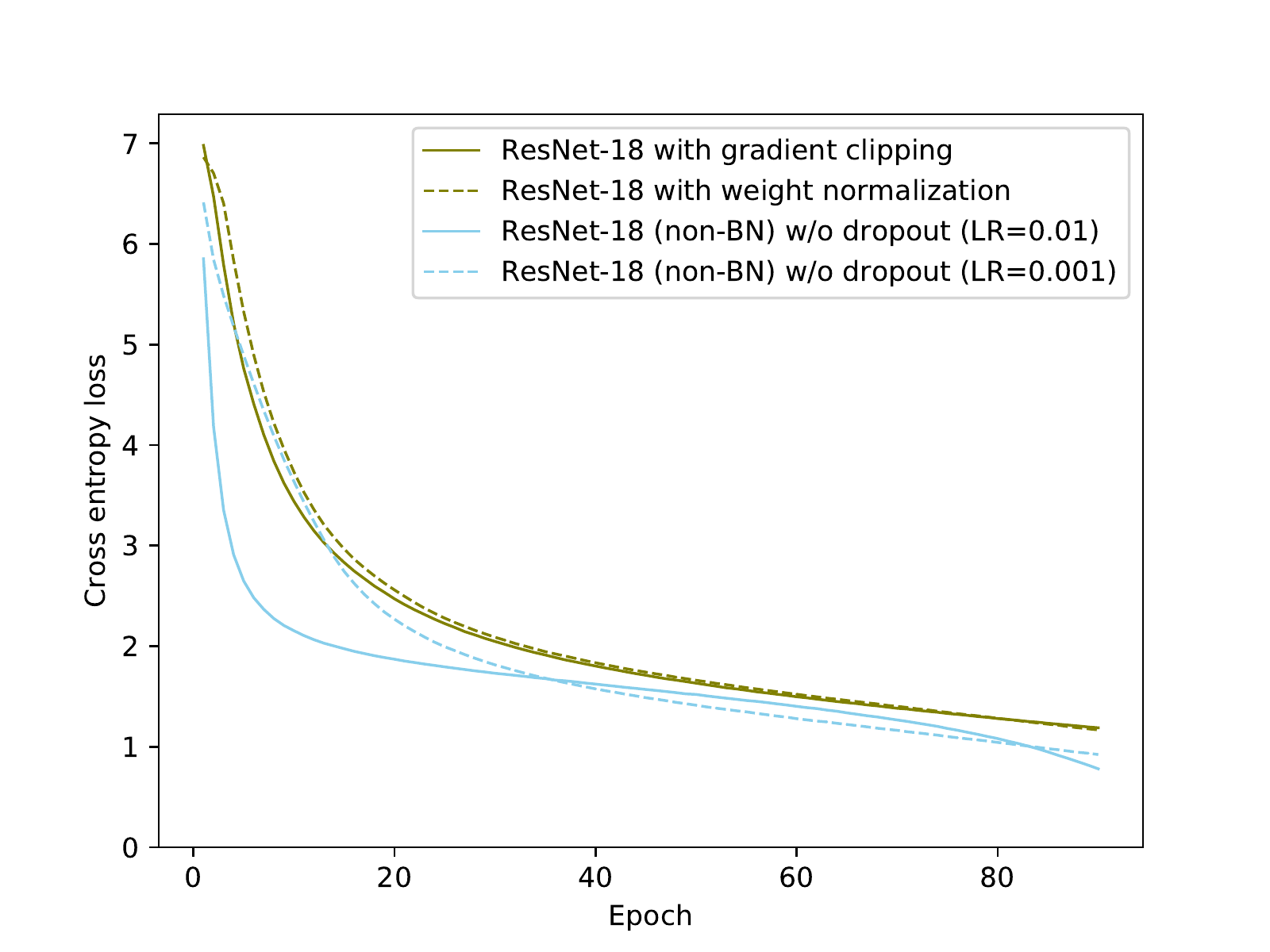}
		\caption{training loss}
		\label{fig:all_non_BN_training_loss}
	\end{subfigure}
	\hfill
	\begin{subfigure}{.5\textwidth}
		\centering
		\includegraphics[scale=0.4]{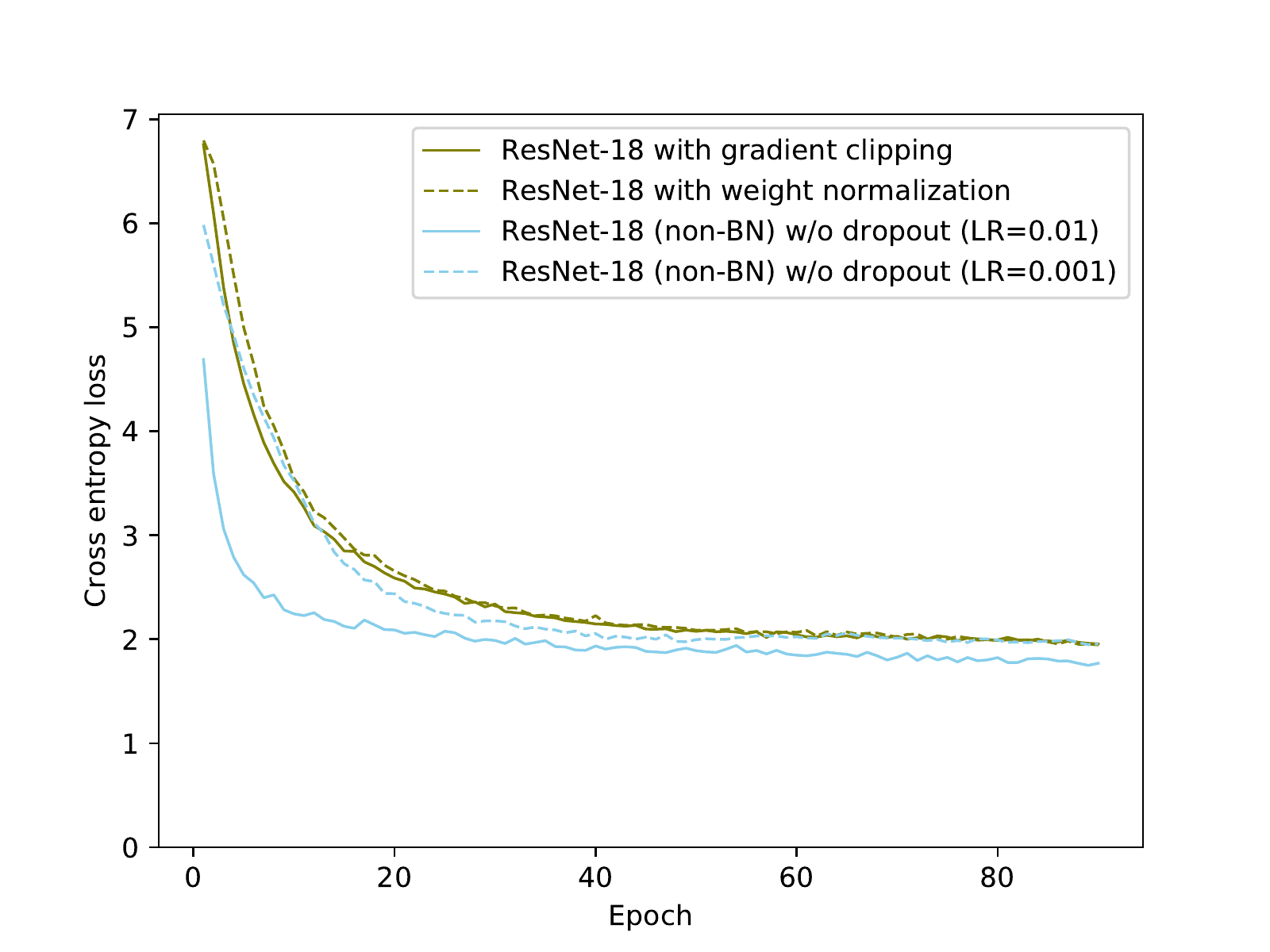}
		\caption{validation loss}
		\label{fig:all_non_BN_validation_loss}
	\end{subfigure}
	\caption{Comparison of non-BN networks trained with only clip norm, only weight norm and with a combination of both.}
	\label{fig:all_non_BN}
\end{figure}

\autoref{fig:all_non_BN}, contains the graphs corresponding to the non-BN methods mentioned in \autoref{tab:Comparison Table for appendix(ResNet-18)},and compares them to the initial non-BN networks. This comparison is basically to highlight the difference between using gradient clipping, weight normalization separately and there performance when these techniques are combined. From \autoref{fig:all_non_BN_training_accuracy}, \autoref{fig:all_non_BN_validation_accuracy} it is evident that although all there is no marked difference in the performance during training phase, validation phase makes the difference. Best results among the discussed   variants is the one with \textit{weight normalization} and \textit{adaptive gradient clipping} with \textit{initial LR} value of \textit{0.01}. Similar behavior can be observed in the \autoref{fig:all_non_BN_validation_loss}, as loss value of initial epochs is relatively and significantly low.

\begin{figure}
	\begin{subfigure}{.5\textwidth}
		\centering
		\includegraphics[scale=0.4]{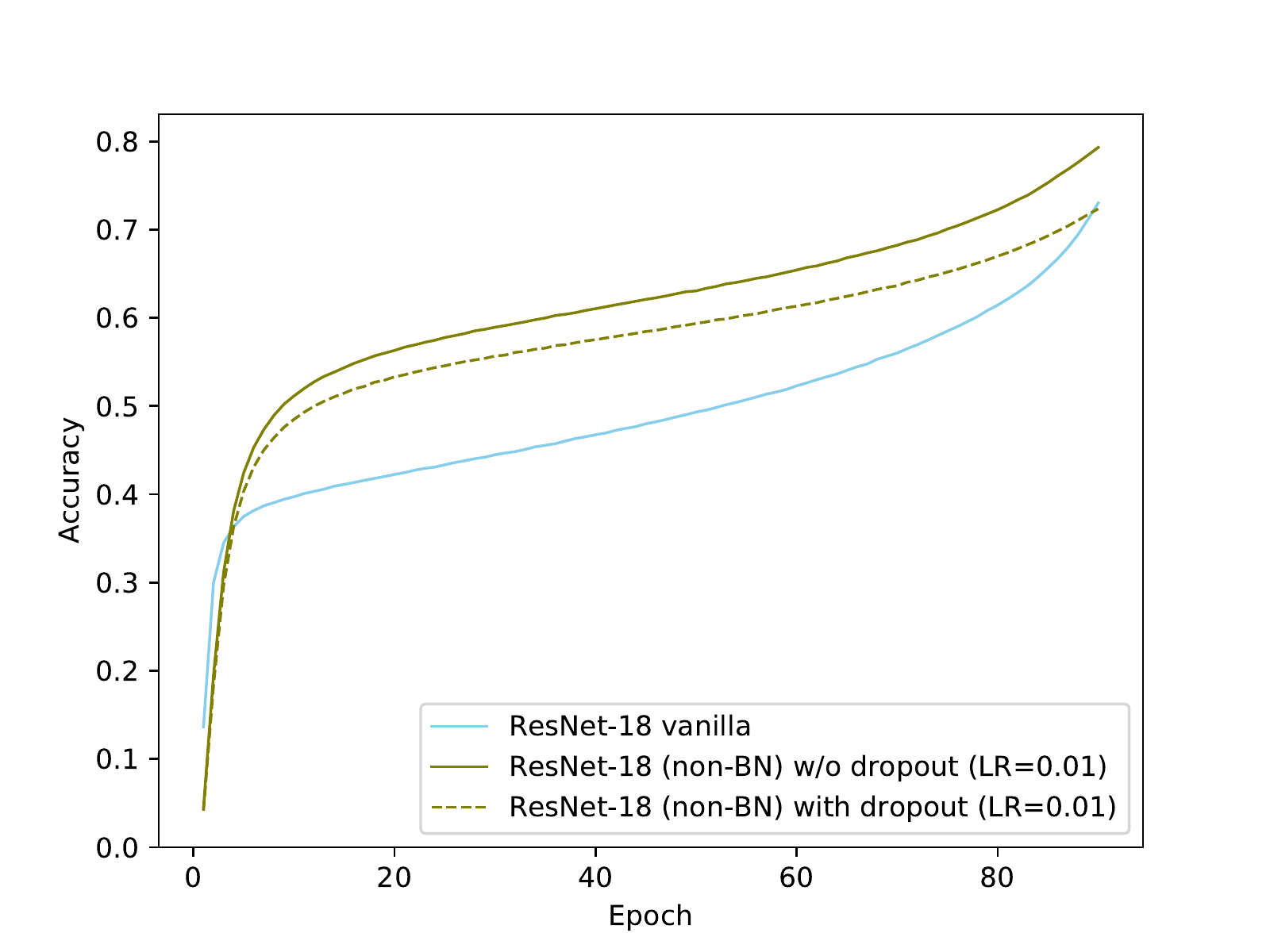}
		\caption{training accuracy}
		\label{fig:with_without_dropout_training_accuracy}
	\end{subfigure}%
	\hfill
	\begin{subfigure}{.5\textwidth}
		\centering
		\includegraphics[scale=0.4]{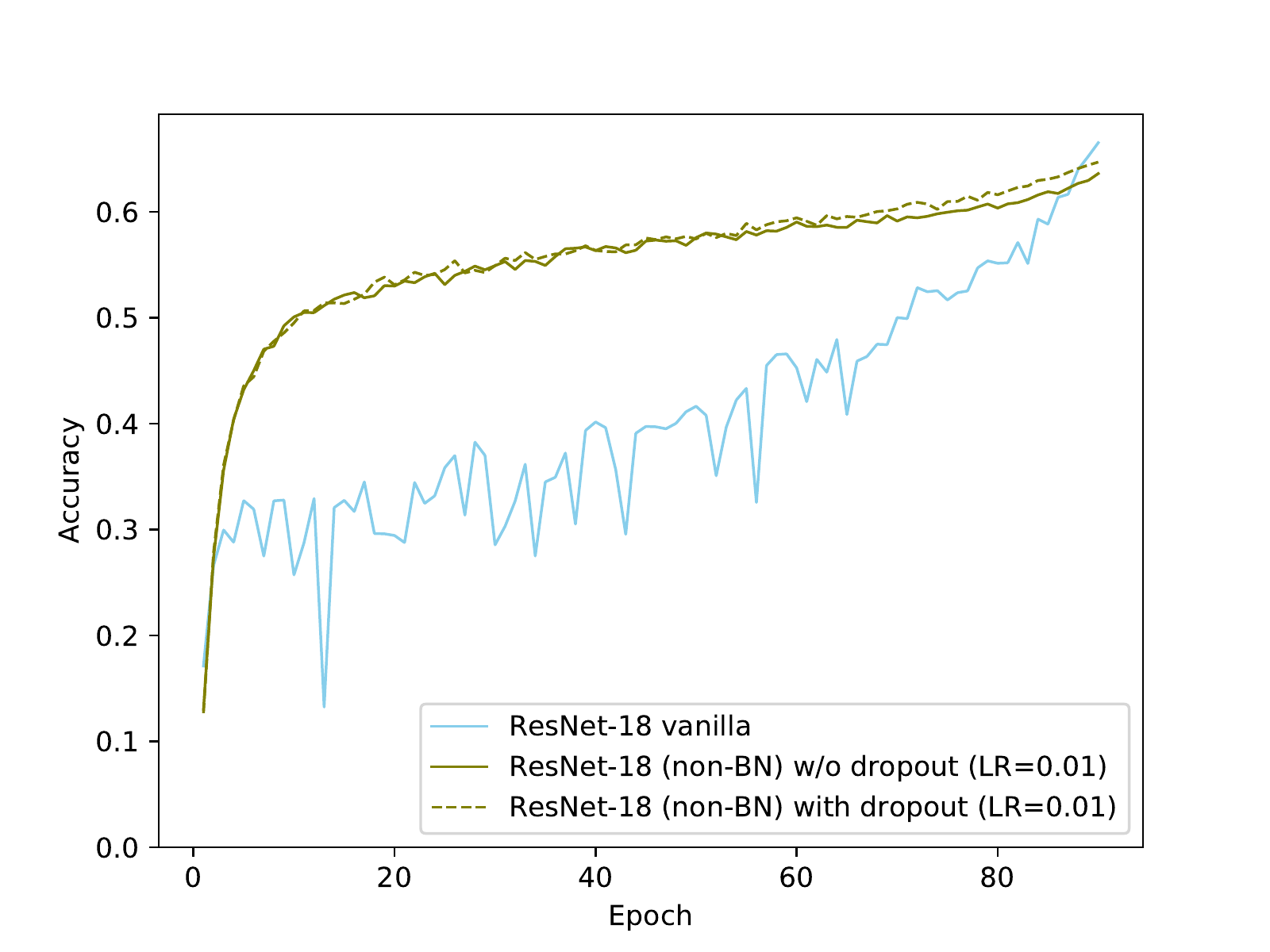}
		\caption{validation accuracy}
		\label{fig:with_without_dropout_validation_accuracy}
	\end{subfigure}
	\begin{subfigure}{.5\textwidth}
		\centering
		\includegraphics[scale=0.4]{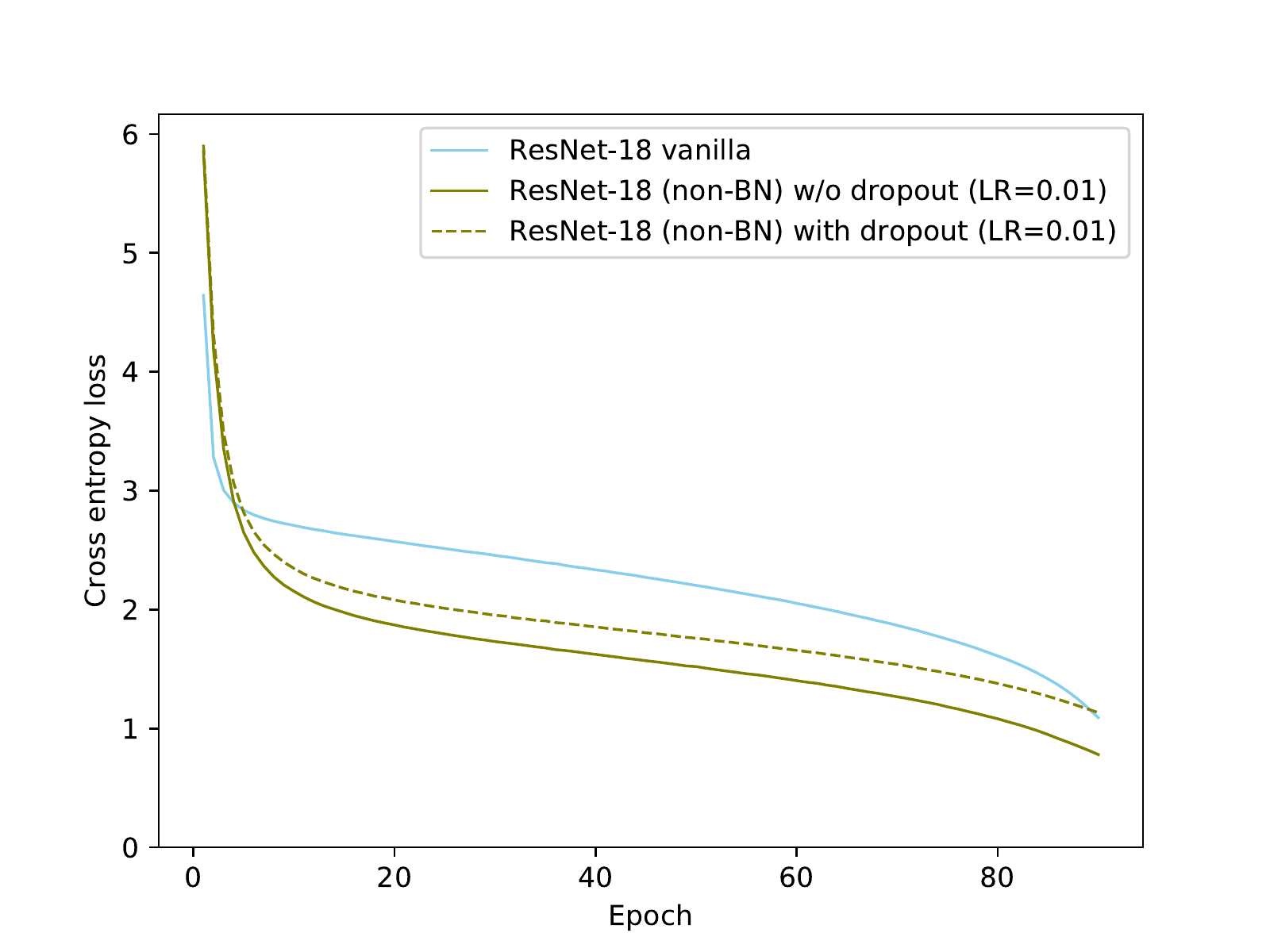}
		\caption{training loss}
		\label{fig:with_without_dropout_training_loss}
	\end{subfigure}
	\begin{subfigure}{.5\textwidth}
		\centering
		\includegraphics[scale=0.4]{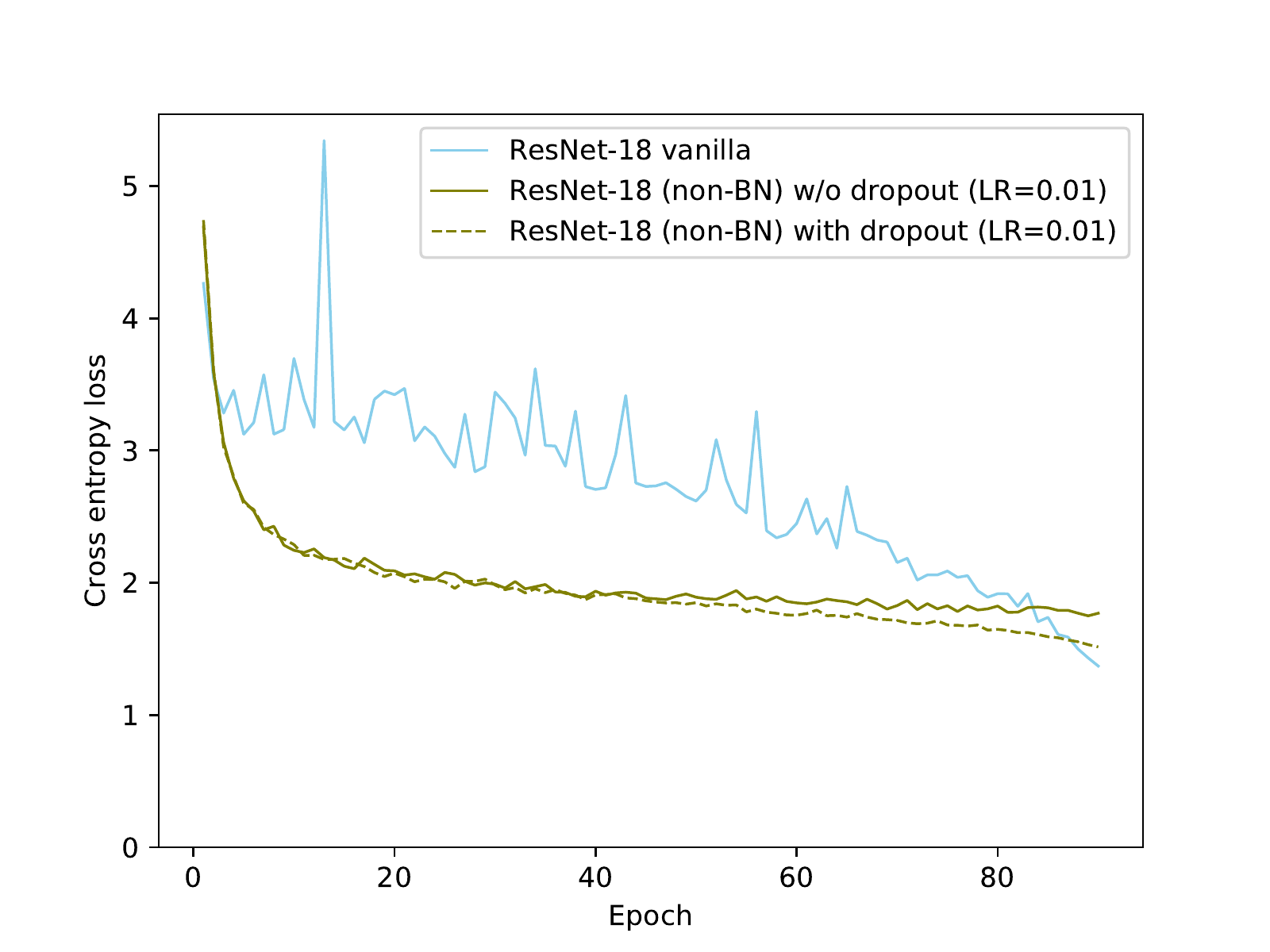}
		\caption{validation loss}
		\label{fig:with_without_dropout_validation_loss}
	\end{subfigure}
	\caption{Comparison of BN and non-BN networks trained with adaptive gradient clipping, weight norm and with/without dropout (ResNet-18).}
	\label{fig:with_without_dropout}
\end{figure}

After arriving at the conclusion that non-BN network works best when it has combination of both \textit{weight normalization} and \textit{adaptive gradient clipping}, especially at a higher \textit{learning rate} (0.01). This knowledge was extended to experiment further and \textit{dropout} layers were included in the architecture to help the regularization of network, preventing it from over-fitting. The values are present in \autoref{tab:Comparison Table for appendix(ResNet-18)}, it can be seen in \autoref{fig:with_without_dropout} network with dropout works better in both training and validation phase unlike the one without (w/o) dropout which over-fits while training but under-performs during the validation. To further improve entire learning process non-BN network mentioned in \autoref{tab:Comparison Table for ResNet-18} is implemented, resulting in a method with promising output.

\begin{figure} 
	\begin{subfigure}{.5\textwidth}
		\centering
		\includegraphics[scale=0.4]{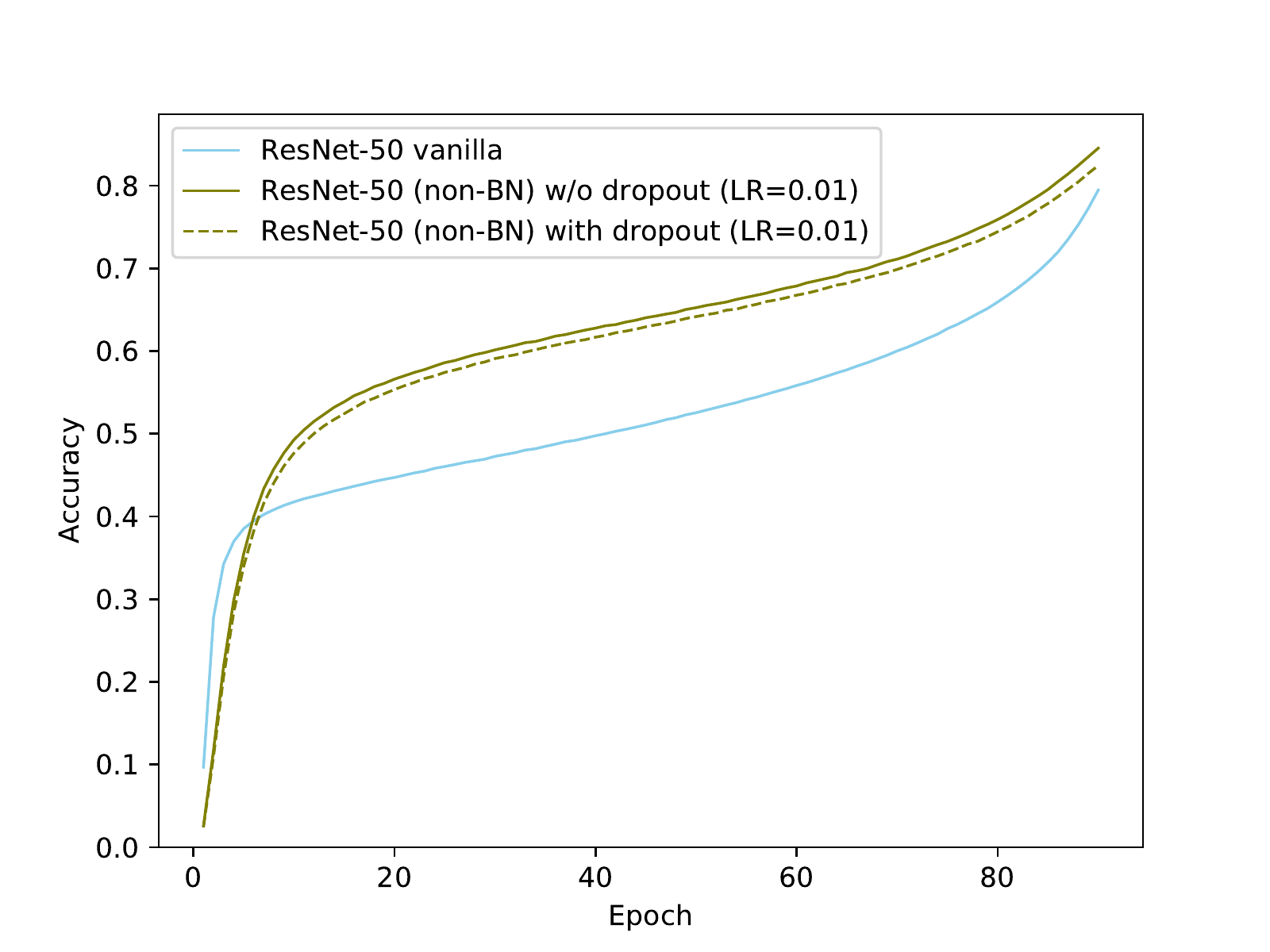}
		\caption{training accuracy}
		\label{fig:with_without_dropout_training_accuracy_resnet50}
	\end{subfigure}%
	\hfill
	\begin{subfigure}{.5\textwidth}
		\centering
		\includegraphics[scale=0.4]{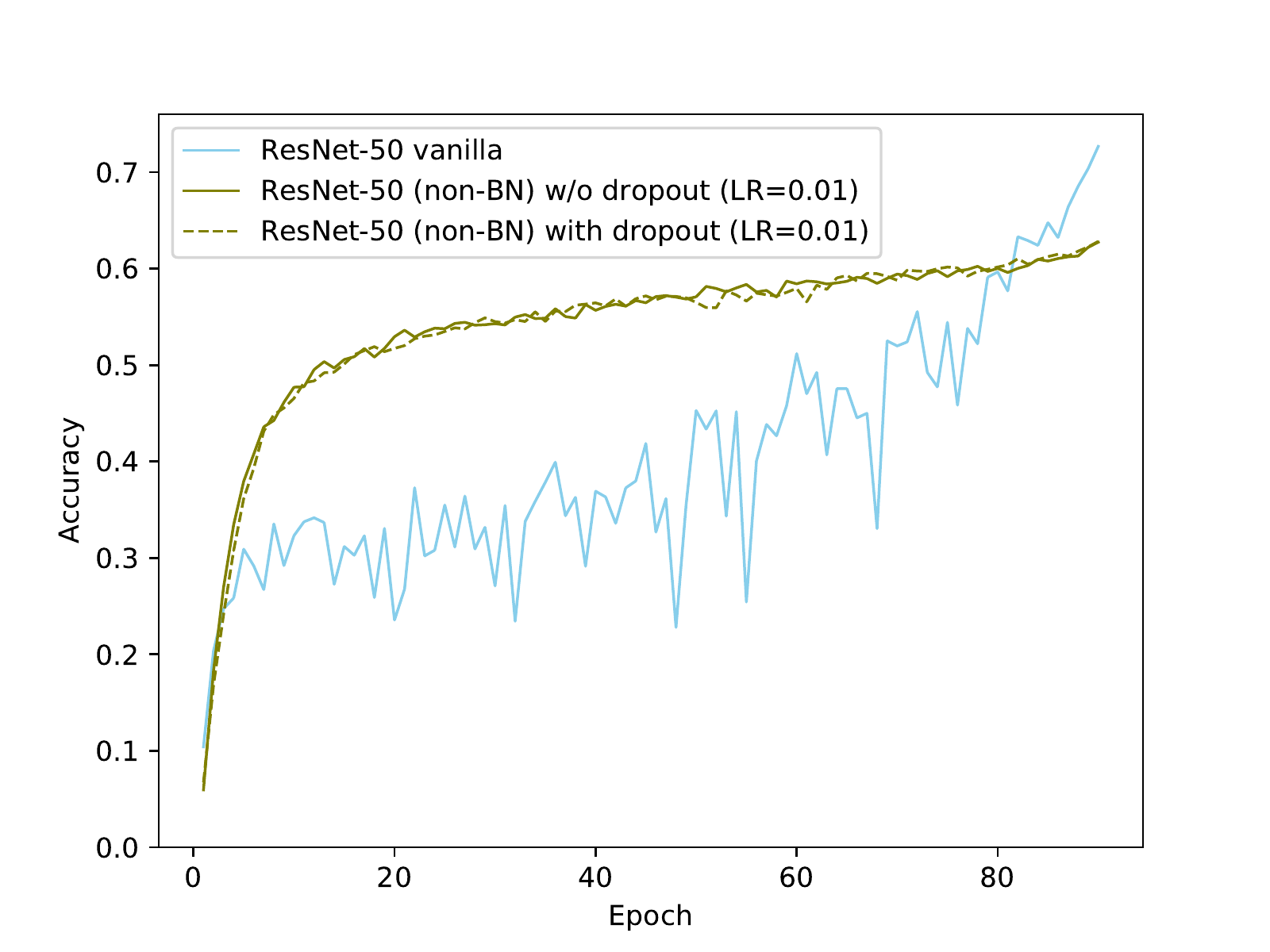}
		\caption{validation accuracy}
		\label{fig:with_without_dropout_validation_accuracy_resnet50}
	\end{subfigure}
	\begin{subfigure}{.5\textwidth}
		\centering
		\includegraphics[scale=0.4]{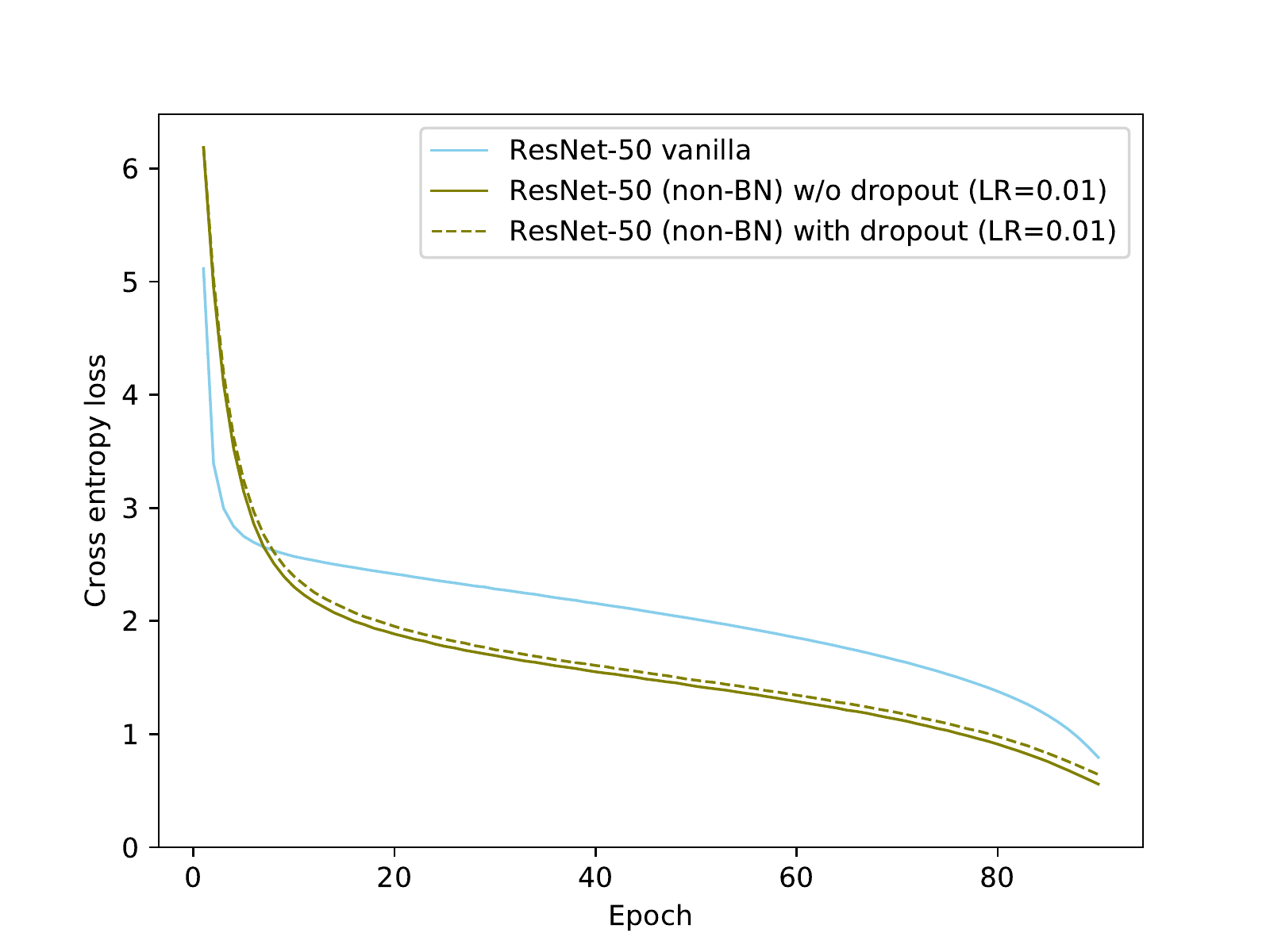}
		\caption{training loss}
		\label{fig:with_without_dropout_training_loss_resnet50}
	\end{subfigure}
	\begin{subfigure}{.5\textwidth}
		\centering
		\includegraphics[scale=0.4]{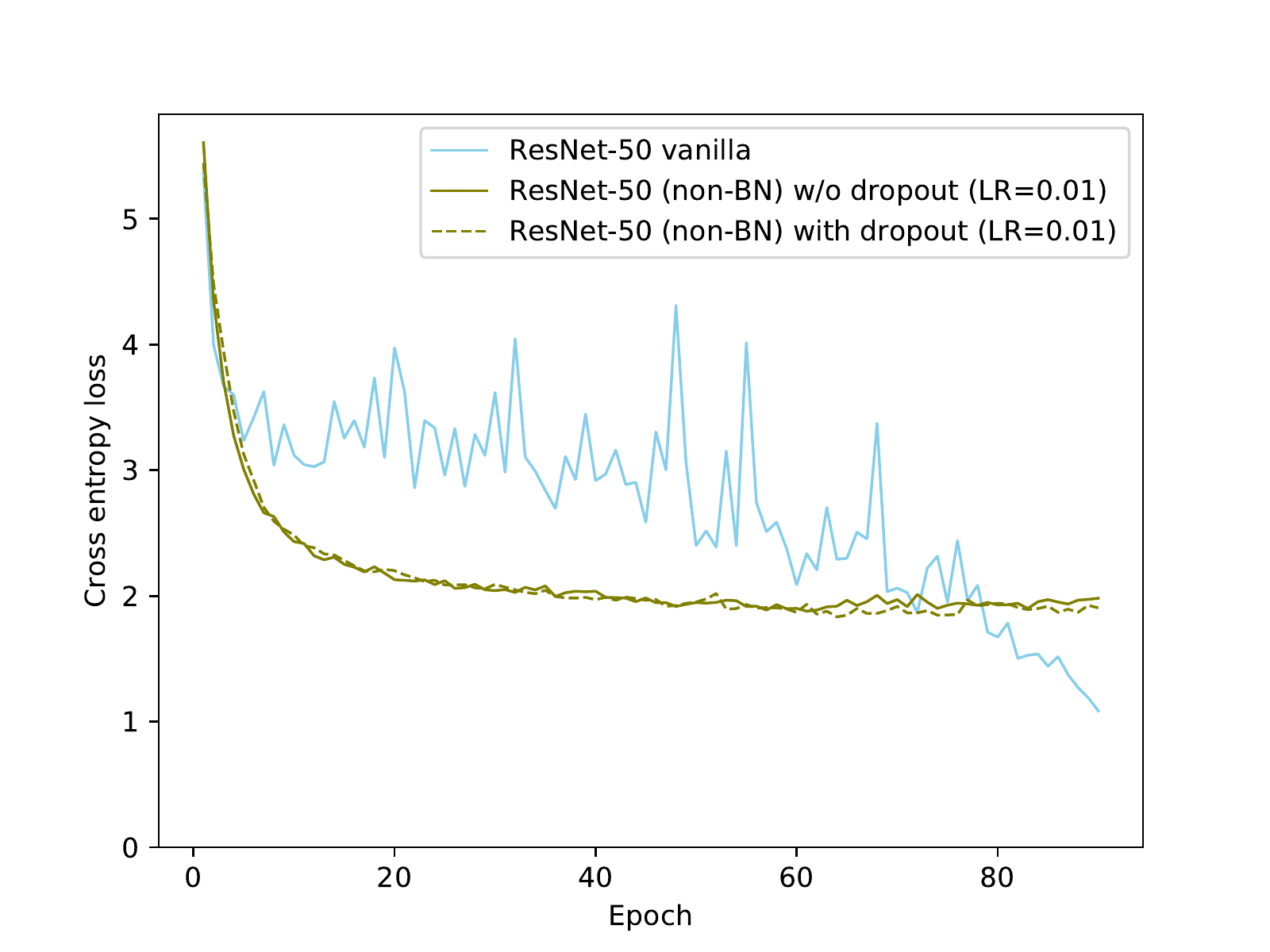}
		\caption{validation loss}
		\label{fig:with_without_dropout_validation_loss_resnet50}
	\end{subfigure}
	\caption{Comparison of BN and non-BN networks trained with adaptive gradient clipping, weight norm and with/without dropout (ResNet-50).}
	\label{fig:with_without_dropout_resnet50}
\end{figure}

\autoref{fig:with_without_dropout_resnet50} shows the graphical representation of results shown in \autoref{tab:Comparison Table for ResNet-50}.Even though the methods which worked best for ResNet-18 were chosen while implementing non-BN versions of ResNet-50, the methods did not work as is evident from the training and validation curve. The training curves show over-fitting, whereas the validation curves show the stagnation around epoch 70, which continues till the entire learning process is complete.  

\end{document}